\documentclass[journal]{IEEEtran}
\ifCLASSINFOpdf
\else
\fi
\usepackage{cite}
   \usepackage[pdftex]{graphicx}
 \usepackage[cmex10]{amsmath}
 \usepackage{algorithmic}
 \usepackage{array}
 \usepackage{url}

\usepackage{amsfonts,amssymb,amsbsy}
\usepackage{amsthm,mathrsfs}
\usepackage{graphicx,color}
\usepackage{verbatim}
\usepackage{pslatex}
\usepackage{cite,url,bm}
\usepackage{algorithm}
\usepackage{algorithmic}
\usepackage{epsf,psfig}
\usepackage{subcaption}
\usepackage{latexsym,amsmath,epsfig}
\usepackage{multirow}
\usepackage{hyperref}
\renewcommand{\algorithmicrequire}{\textbf{Input:}}
\renewcommand{\algorithmicensure}{\textbf{Output:}}

\newcommand{\vect}[1]{\pmb{#1}}
\newcommand{\mat}[1]{\pmb{#1}}

\def\ben{\begin{equation*}}
\def\een{\end{equation*}}
\def\be{\begin{equation}}
\def\ee{\end{equation}}
\def\beaa{\begin{eqnarray*}}
\def\eeaa{\end{eqnarray*}}
\def\bea{\begin{eqnarray}}
\def\eea{\end{eqnarray}}
\def\tb{\textbf}
\def\tcb{\textcolor{blue}}
\def\bleq{\begin{flalign}}
\def\eleq{\end{flalign}}
\DeclareMathOperator{\Tr}{Tr}

\begin{document}
%
\title{Sparsity-based Color Image Super Resolution via Exploiting Cross Channel Constraints}
%
%
%

\author{Hojjat S. Mousavi,~\IEEEmembership{Student Member,~IEEE,}
        and Vishal Monga,~\IEEEmembership{Senior Member,~IEEE,}
\thanks{H. S. Mousavi and V. Monga are with the Department
of Electrical Engineering, The Pennsylvania State University, University Park, PA, 16802, USA e-mail: \href{mailto:hojjat@psu.edu}{hojjat@psu.edu}}
\thanks{Research is supported by NSF CAREER award to V.M.}
\thanks{Manuscript submitted September, 2016; }}

%
%

\markboth{Submitted to Transactions on Image Processing,~Vol.~, No.~, September~2016}%
{Mousavi \MakeLowercase{\textit{et al.}}: Color Image Super Resolution via Joint Sparse Representations and Learning}
%



\maketitle

\begin{abstract}
Sparsity constrained single image super-resolution (SR) has been of much recent interest. A typical approach involves sparsely representing patches in a low-resolution (LR) input image via a dictionary of example LR patches, and then using the coefficients of this representation to generate the high-resolution (HR) output via an analogous HR dictionary. However, most existing sparse representation methods for super resolution focus on the luminance channel information and do not capture interactions between color channels. In this work, we extend sparsity based super-resolution to multiple color channels by taking color information into account. Edge similarities amongst RGB color bands are exploited as cross channel correlation constraints. These additional constraints lead to a new optimization problem  which is not easily solvable; however, a tractable solution is proposed to solve it efficiently. Moreover, to fully exploit the complementary information among color channels, a dictionary learning method is also proposed specifically to learn color dictionaries that encourage edge similarities. Merits of the proposed method over state of the art are demonstrated  both visually and quantitatively using image quality metrics.
\end{abstract}


%
\IEEEpeerreviewmaketitle

\section{Introduction}
%
%
%
%
\label{sec:intro}
\IEEEPARstart{S}{uperresolution}  is a  branch of image reconstruction and an active area of research that focuses on the enhancement of image resolution. Conventional Super-Resolution (SR) approaches require multiple Low Resolution (LR) images of the same scene as input and maps them to a High Resolution (HR) image based on some reasonable assumptions, prior knowledge, or capturing the diversity in LR images \cite{Freeman:ExampleBasedSR_CompGraph2002 ,Farsiu:MultiFrameSR_TIP2004, Park:ReviewSR_SPM2003}. This can be seen as an inverse problem of recovering the high resolution image by fusing the low resolution images of the scene. The recovered image should produce the same low resolution images if the physical image formation model is applied to the HR image. However, SR task is a severely  ill-posed problem since much information is lost in the process of going from high resolution images to low resolution images and hence the solution is not unique. Consequently, strong prior information is incorporated to yield realistic and robust solutions. Example priors include knowledge of the underlying scene, distribution of pixels, historical data, smoothness and edge information and so on so forth. \cite{Tappen:SparsePriorSR_2003,Fattal:SRstatistic_ACM2007,Dai:edgeSR_CVPR2007, Minaee:Segmen_ICIP2015}

In contrast to conventional super resolution problem with multiple low resolution images as input, single image super-resolution methods have been developed recently that generate the high resolution image only based on a \emph{single} low resolution image. Classically, solution to this problem is based on example-based methods exploiting nearest neighbor estimations, where pairs of low and high resolution image patches are collected and each low resolution patch is mapped to a corresponding high resolution patch. Freeman \emph{et al.} \cite{Freeman:ExampleBasedSR_CompGraph2002} proposed an estimation scheme where high-frequency details are obtained by taking nearest neighbor based estimation on low resolution patches. Glasner \emph{et al.} \cite{Glasner:SRSingle2009ICCV} used the observation that patches in a natural image tend to redundantly recur many times inside the image, both within the same scale, as well as across different scales and approached the single image super resolution problem. An alternate mapping scheme was proposed by Kim \emph{et al.} \cite{Kim:SRNaturalPrior_PAMI2010} using kernel ridge regression.

Many learning techniques have been developed which attempt to capture the co-occurrence of low resolution and high resolution image patches. \cite{Sun:HallucinationSR_CVPR2003} proposes a Bayesian approach  by using Primal Sketch priors. Inspired by manifold forming methods like locally linear embedding (LLE), Chang \emph{et al.} \cite{Chang:NeighborEmbeddingSR_CVPR2004} propose a neighbourhood embedding approach. Specifically, small image patches in the low and high resolution images form manifolds with similar local geometry in two distinct feature spaces and local geometry information is used to reconstruct a patch using its neighbors in the feature space.


More recently, sparse representation based methods have been applied to the single image super resolution problem. Essentially in these techniques, a historical record of typical geometrical structures observed in images is exploited and examples of high and low resolution image patches are collected as dictionary (matrix). Yang \emph{et al.} proposed to apply sparse coding for retrieving the high resolution image from the LR image \cite{YangAndWright:SparseSR_TIP2010}. Zeyde \emph{et al.} extended this method to develop a local Sparse-Land model on image patches \cite{Zeyde:SR_Springer2012}. Timofte \emph{et al.} proposed the Anchored Neighborhood Regression (ANR) method which uses learned dictionaries in combination with neighbor embedding methods \cite{Timofte:AnchoredANR_ICCV2013, Timofte:AnchoredARN+_ACCV2014}. Other super resolution methods based on statistical signal processing or dictionary learning methods have been proposed by \cite{Peleg:StatisticalSR_2014TIP, Zhou:KPCAcodingSR2015_SPL, Polatkan:BayesianSR_2015PAMI, Huang:TransformedExampleSR_2015cvpr, Zhang:MultiLinearMap_2015TIP, He:BetaProcessDL_CVPR2013}.

%

On top of sparsity based methods, learning based methods have also been exploited for SR problem to learn dictionaries that are more suitable for this task. Mostly, dictionary learning or example-based learning methods in super-resolution use an image patch or feature-based approach to learn the relationship between high resolution scenes and their low resolution counterparts.  Yang \emph{et al.} \cite{YangAndWright:RawSparseSR_CVPR2008} propose to use collection of raw image patches as dictionary elements in their framework. Subsequently, a method that learns LR and HR dictionaries jointly was proposed in \cite{YangAndWright:SparseSR_TIP2010}. A semi-coupled dictionary learning (SCDL) model and a mapping function was proposed in  \cite{Wang:SemiCoupledDL_CVPR2012} where the learned dictionary pairs can characterize the structural features of the two image domains, while the mapping function reveals the intrinsic relationship between the two. In addition, coupled dictionary learning for the same problem was proposed in \cite{Yang:CoupledDicLearnSR_TIP2012}, where the learning process is modeled as a bilevel optimization problem. Dual or joint filter learning in addition to dual (joint)  dictionaries was developed by Zhang \emph{et al.} \cite{Zhang:DualDLSR_ICME2011}.

\subsection{Sparsity Based Single Image Super-Resolution}
In the setting proposed by Yang \emph{et al.} (ScSR) \cite{YangAndWright:SparseSR_TIP2010} a large collection of corresponding high resolution and low resolution image patches is obtained from training data. In this framework, the low resolution information can either be in the form of raw image patches, high frequency or edge information, or any other types of representative features, while high resolution information is in the form of image pixels to ensure reconstruction of high resolution images. Using methods mentioned for dictionary learning in SR task and sparsity constraints, high resolution and low resolution dictionaries are jointly learned such that they are capable of representing the LR image patches and their corresponding HR counterparts using the same sparse code. Once the dictionaries are learned, the algorithm searches for a sparse linear representation of each patch of LR image based on the following sparse coding optimization:
\bea
     \vect x^\ast  = \arg\min_{\vect x} ~~ \frac{1}{2}||\vect y_l - \mat D_l \vect x||_2^2 + \lambda || \vect x||_1 \label{Eq:ScSRopt}
\eea
where $\mat D_l$ is the learned low resolution dictionary (or dictionary that is learned based on features extracted from LR patches), $\vect x$ is the   sparse code representing the LR patch (or features extracted from LR patch) with respect to $\mat D_l$ and $\lambda$ is a regularizer parameter for enforcing the sparsity prior and regularizing the ill-posed problem. This is the familiar and famous LASSO \cite{Sprechmann:CHI-LASSO_TSP2011, Mousavi:ICR_SPL2015} problem which can be easily solved using any sparse solver toolbox. The high resolution reconstruction ($\vect y_h$) of each low resolution patch or features of the patch ($\vect y_l$) is then reconstructed using the same sparse code according to the HR dictionary as: $\vect y_h = \mat D_h \vect x^\ast $. Joint dictionary learning for SR considers the problem of learning two joint dictionaries $\mat D_l$ and $\mat D_h$ for two features spaces (low resolution and high resolution domains) which are assumed to be tied by a certain mapping function \cite{YangAndWright:SparseSR_TIP2010, Wang:SemiCoupledDL_CVPR2012}. The assumption is that $\vect x$, the sparse representation of $\vect y_l$ based on learned low resolution dictionary, should be  the same as that of $\vect y_h$ according to $\mat D_h$. The following optimization problem encourages this idea and learns low resolution and high resolution image dictionaries according to the same sparse code:
\bea
    \min_{\mat D_l, \mat D_h,\{\vect x^i\}}  \frac{1}{N} \sum_{i=1}^{N} \frac{1}{2}\| \vect y_l^i - \mat D_l \vect x^i \|_2^2 + \frac{1}{2}\| \vect y_h^i - \mat D_h \vect x^i \|_2^2 + \lambda \|\vect x^i\|_1  \nonumber \\
    \text{st.~~} \|\mat D_l(:,k) \|_2^2 \le 1, ~ \|\mat D_h(:,k) \|_2^2 \le 1, ~~k=1,2,...,K.
\eea
where $N$ is the number of training sample pairs and $K$ is the number of desired dictionary basis atoms. $\mat D(:,k)$ denotes the $k^{th}$ column of the matrix $\mat D$.

\subsection{Motivation and Contributions}
Most of super-resolution methods, especially in single image SR literature, have been designed to increase the resolution of a single channel (monochromatic) image. A related yet more challenging problem, color super-resolution, addresses enhancing resolution of color (multi-channel) low resolution images to increase their spatial resolution.
The typical solution for color super resolution involves applying SR algorithms to each of the color channels independently \cite{Shah:ColorVideoSR_TIP1999, Tom:ColorVideoSR_TIP2001}. Another approach which is more common is to transform the problem to a different color space such as YCbCr, where chrominance information is separated from luminance, and SR is applied only to the luminance channel \cite{Yang:CoupledDicLearnSR_TIP2012, YangAndWright:SparseSR_TIP2010, Timofte:AnchoredANR_ICCV2013} since the human eye is more sensitive to luminance information than chrominance information. Both of these methods are suboptimal for color super-resolution problem as they do not fully exploit the complementary information that may exist in different color channels. Moreover, the correlation across the color bands and the cross channel information are ignored in these ways of handling color super-resolution. In addition to this, many images have more information in the color channels rather than only luminance channel. For instance, Fig. \ref{Fig:Motivation} illustrates a synthetic image where there is much more color information (prominent edges) in chrominance channels (Cb and Cr) than luminance channel (Y).
In traditional multi-frame super resolution problem, color information has been used in different ways to enhance super resolution results. Farsiu \emph{et al.} \cite{Farsiu:ColorDemosaicSR_TIP2006} proposed a multi-frame demosaicing and super resolution framework for color images using different color regularizers. Belekos \emph{et al.} proposed multi channel video super resolution in \cite{Belekos:MultiChannelVideoSR_2010TIP} and general color dictionary learning for image restoration is proposed in \cite{Mairal:SparsityImageRestoration_2008TIP, Xu:QuaternionColorSR_2015TIP, Jung:MumfordShahSR_2011TIP}. Other methods that use color channel information are proposed in \cite{Shen:SparsityColorSR_2013AppliedMechanics, Dai:SoftCutColorSR_2009TIP, Gong:NeighborEmbedColorSR_2011CompInfo, Wang:EdgeColorSR_2015APPC, Maalouf:GroupletColorSR_2009EuroSP, Liu:ColorizationForSR_2010ECCV, Cheng:RGBSR_2015IETImaPrc, Maalouf:GroupSR_2012IETImgPrc}.

We develop a sparsity based Multi-Channel (i.e.\ color) constrained Super Resolution (MCcSR) framework. \textbf{The key contributions of our work}\footnote{{Preliminary version of work was presented in ICIP conference 2016 \cite{Mousavi:ColorSR_2016ICIP}}} are as follows:
\begin{itemize}
    \item We explicitly address the problem of color image super-resolution by inclusion of color regularizers in the sparse coding for SR. These color regularizers capture the cross channel correlation information existing in different color channels and exploit it to better reconstruct super-resolution patches. The resulting optimization problem with added color-channel regularizers is not easily solvable and a tractable solution is proposed.
    \item The amount of color information is not the same in each region of the image and in order to be able to force color constraints we develop a measure that captures the amount of color information and then use it to balance the effect of color regularizers. Therefore, an adaptive color patch processing scheme is also proposed in the paper where patches with stronger edge similarities are optimized with more emphasis on the color constraints.
    \item In most dictionary learning algorithms for super-resolution, only the correspondence between low and high resolution patches is considered. However, we propose to learn dictionaries whose atoms (columns) are not only low resolution and high resolution counterparts of each other, but also in the high resolution dictionary in particular, we incorporate color regularizers such that the resulting learned high resolution patches exhibit high edge correlation across RGB color bands. 
    \item {\em Reproducibility:} All results in this paper are completely reproducible. The MATLAB code as well as images corresponding to the SR results are made available at: \tcb{\url{http://signal.ee.psu.edu/MCcSR.html}}.
\end{itemize}

The rest of this paper is organized as follows: In Section \ref{Sec:ColorSR}, we generalize the sparsity-based super resolution framework to multiple (color) channels  and motivate the choice of color regularizers. These color regularizers are used in Section \ref{Sec:ColorDL} to assist learning of color adaptive dictionaries suitable for color super resolution task. Section \ref{Sec:Experiments} includes experimental validation which demonstrates the effectiveness of our approach by comparing it with state-of-the-art image SR techniques. Concluding remarks are collected in Section \ref{Sec:Conclusion}.

\section{Sparsity Constrained Color Image Super Resolution}
\label{Sec:ColorSR}

\subsection{Problem formulation}
\label{sec:formulation}

\begin{figure}
  \centering
  \includegraphics[width=\columnwidth]{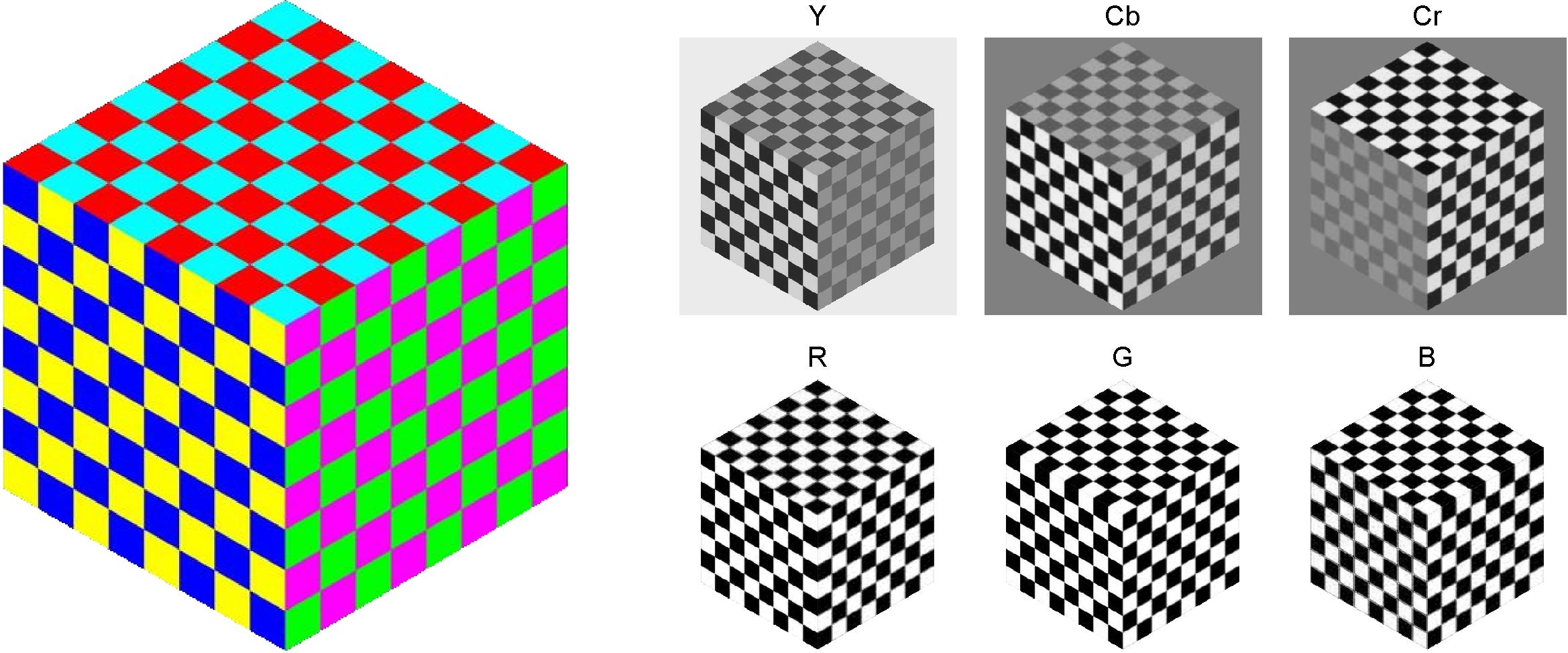}\\
  \caption{Color chessboard cube and  color channel components. }\label{Fig:Motivation}
\end{figure}

A characteristic associated with most natural images is strong correlation between high-frequency spatial components across the color (RGB) channels. This is based on the intuition that a luminance edge for example is spread across the RGB channels \cite{Farsiu:ColorDemosaicSR_TIP2006, Srinivas:ColorSR_CIC2011}. Fig. \ref{Fig:Motivation} illustrates this idea.

\begin{figure}
  \centering
  \centering
  \includegraphics[width=0.9\columnwidth]{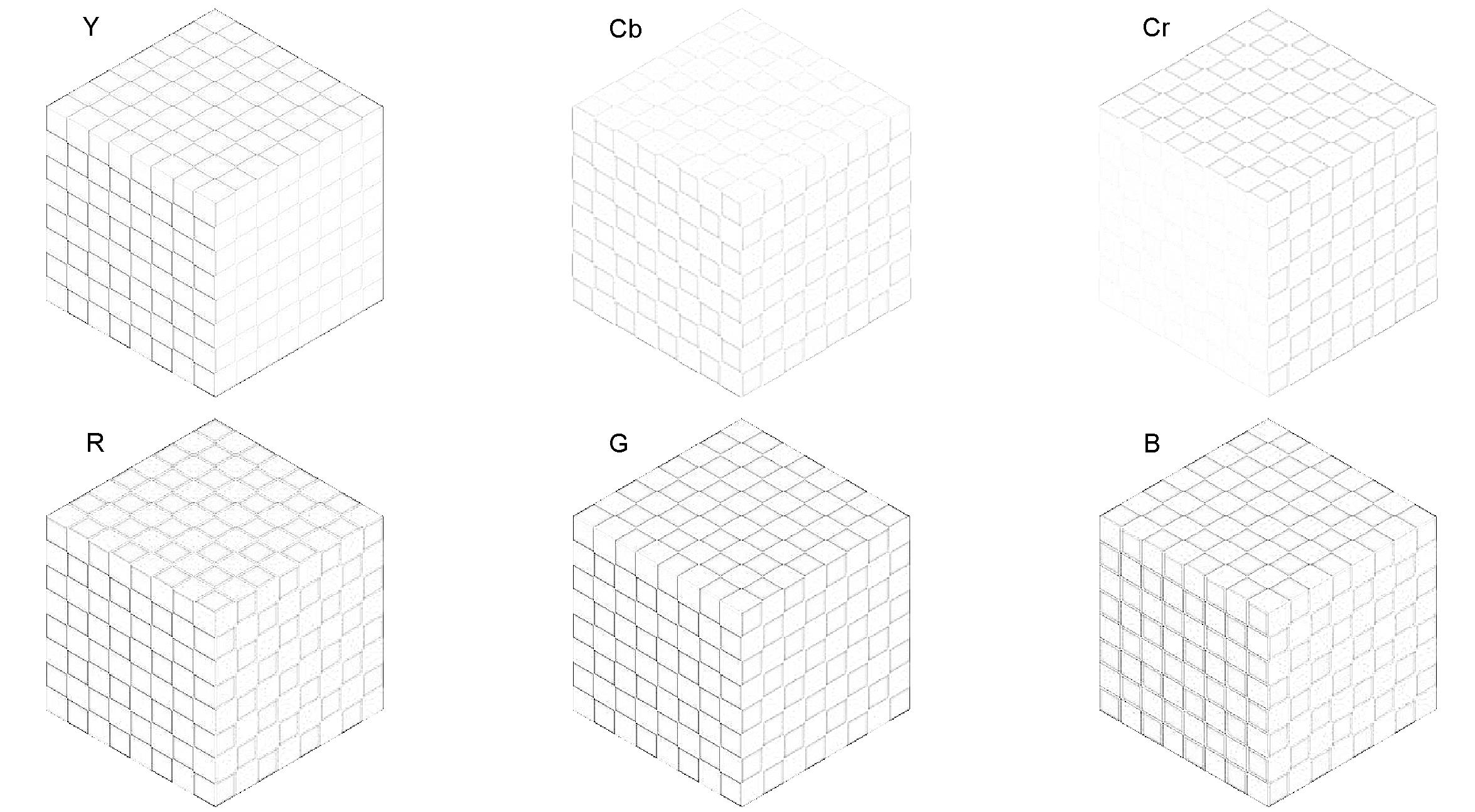}\\
  \caption{Edges for color channels of chessboard cube. }\label{Fig:MotivationEdge}
\end{figure}

 We can hence encourage the edges across color channels to be similar to each other. Fig. \ref{Fig:MotivationEdge} also shows that RGB edges are far more close to each other than YCbCr edges. Such ideas have been exploited in traditional image fusion type super-resolution techniques \cite{Farsiu:ColorDemosaicSR_TIP2006}, yet sparsity-based single image super resolution lacks a concrete color super resolution framework.
Edge similarities across RGB color channels may be enforced in the following manner  \cite{Farsiu:ColorDemosaicSR_TIP2006, Keren:ColorSR_1999MachineVision, Menon:ColorDemosaick_2009TIP}.
\bea
        \|\mat S_\mu \vect y_{h_\mu} - \mat S_\nu \vect y_{h_\nu}\|_2 < \epsilon_{\mu\nu} &,~ \mu,\nu\in\{r,g,b\} &,~ \mu\ne\nu \label{Eq:CrossConstrains}
\eea
where $r,g$ and $b$ subscripts are indicating signals in R, G and B channels and $\mat S$ matrix is a high-pass edge detector filter as in \cite{Srinivas:ColorSR_CIC2011}. For instance, $\mat S_r \vect y_{h_r}$  illustrate the edges in red channel of the desired high resolution image. These constraints are essentially enforcing the edge information across color channels to be similar in {\em high resolution patches}. The underlying assumption here is that the high resolution patches need  to be known beforehand which is not true in practice. We recognize however that these constraints can be equivalently posed on the sparse coefficient vector(s) corresponding to the individual color channels, since: $\vect y_{h_r} = \mat D_{h_r} \vect x_r,~~~\vect y_{h_g} = \mat D_{h_g} \vect x_g,~~~\vect y_{h_b} = \mat D_{h_b} \vect x_b$.

Note that sparse codes for different color channels are no longer independent and they may be jointly determined by solving the following optimization problem:
\bea
     [\vect x_r,\vect x_g,\vect x_b]  &=&\arg\min \sum_{c \in \{r,g,b\}} \frac{1}{2} \|\vect y_{l_c} -\mat D_{l_c} \vect x_c\|_2^2 +  \lambda \|\vect x_c \|_1 \nonumber\\
&&     +\tau \Big[ \|\mat S_r \mat D_{h_r} \vect x_r - \mat S_g \mat D_{h_g} \vect x_g \|_2^2 \nonumber\\
&&     ~~+\|\mat S_g \mat D_{h_g} \vect x_g - \mat S_b \mat D_{h_b} \vect x_b\|_2^2 \nonumber\\
&&     ~~+\|\mat S_b \mat D_{h_b} \vect x_b-\mat S_r \mat D_{h_r} \vect x_r\|_2^2 \Big]. \label{Eq:MainOptProb}
\eea
where the cost function is equivalent to the following:
\bea
     L_1&=&   \sum_{c \in \{r,g,b\}} \Big[ \frac{1}{2} \|\vect y_{l_c} -\mat D_{l_c} \vect x_c\|_2^2 +  \lambda \|\vect x_c \|_1  \nonumber\\
     &&+2\tau \vect x_c^T \mat D_{h_c}^T \mat S_c^T \mat S_c \mat D_{h_c} \vect x_c  \Big ]     -2\tau \Big [  \vect x_r^T \mat D_{h_r}^T \mat S_r^T \mat S_g \mat D_{h_g} \vect x_g +   \nonumber\\
     && \vect x_g^T \mat D_{h_g}^T \mat S_g^T \mat S_b \mat D_{h_b} \vect x_b    +   \vect x_b^T \mat D_{h_b}^T \mat S_b^T \mat S_r \mat D_{h_r} \vect x_r  \Big ] \label{Eq:Cost2}
\eea
For simplicity, we assume the same regularization parameters $\tau$ and $\lambda$ for each of the edge difference terms and color channels. The high-pass edge detectors ($\mat S_r,\mat S_g,\mat S_b$) are also chosen to be the same for each color channel. It is worth mentioning that if  $\tau = 0$, \eqref{Eq:MainOptProb} reduces to three independent sparse coding problems (ScSR) for each color channel. With the cross channel regularization terms, these sparse codes are no longer independent and \eqref{Eq:MainOptProb} presents a challenging optimization problem in contrast with the optimization problem corresponding to single channel sparsity based super resolution. In the new problem, the additional color channel regularizers are of quadratic nature and make the optimization problem more challenging to solve. Next, we propose a tractable solution.

\subsection{Solution to the optimization problem}
We introduce the following vectors and matrices:
\beaa
    \vect x =
    \begin{bmatrix}
    \vect x_r  \\
    \vect x_g  \\
    \vect x_b  \\
    \end{bmatrix}_{3m\times 1}
    ,
    \vect y_l =
    \begin{bmatrix}
    \vect y_{l_r}  \\
    \vect y_{l_g}  \\
    \vect y_{l_b}  \\
    \end{bmatrix}_{3p\times 1}
    ,~
    \mat P =
    {\begin{bmatrix}
    \mat 0       & \mat 0       & \mat I      \\
    \mat I       & \mat 0       & \mat 0      \\
    \mat 0       & \mat I       & \mat 0 \\
    \end{bmatrix}}_{3m \times 3m}
\eeaa
\beaa
    \begin{bmatrix}
    \vect x_b  \\
    \vect x_r  \\
    \vect x_g  \\
    \end{bmatrix} =
    \underbrace{\begin{bmatrix}
    \mat 0       & \mat 0       & \mat I      \\
    \mat I       & \mat 0       & \mat 0      \\
    \mat 0       & \mat I       & \mat 0 \\
    \end{bmatrix}}_{\mat P}
    \underbrace{\begin{bmatrix}
    \vect x_r  \\
    \vect x_g  \\
    \vect x_b  \\
    \end{bmatrix}}_{\vect x} = \mat P \vect x
\eeaa
\bea
    \mat D_l =
        \begin{bmatrix}
        \mat D_{l_r} & \mat 0       & \mat 0      \\
        \mat 0       & \mat D_{l_g} & \mat 0      \\
        \mat 0       & \mat 0       & \mat D_{l_b}\\
        \end{bmatrix},~
    \mat D_h =
        \begin{bmatrix}
        \mat D_{h_r} & \mat 0       & \mat 0      \\
        \mat 0       & \mat D_{h_g} & \mat 0      \\
        \mat 0       & \mat 0       & \mat D_{h_b}\\
        \end{bmatrix}
        \label{Eq:DicsDef}
\eea
\beaa
    \mat S =
        \begin{bmatrix}
        \mat S_{r} & \mat 0       & \mat 0      \\
        \mat 0       & \mat S_{g} & \mat 0      \\
        \mat 0       & \mat 0       & \mat S_{b}\\
        \end{bmatrix}_{3p\times 3p},~~
    \mat P_s =
        \begin{bmatrix}
    \mat 0       & \mat 0       & \mat I      \\
    \mat I       & \mat 0       & \mat 0      \\
    \mat 0       & \mat I       & \mat 0 \\
    \end{bmatrix}_{3p\times 3p}
\eeaa

Where $\vect x$ and $\vect y_l$ respectively are concatenation of sparse codes and low resolution image patches (or features) in different color channels. $\mat P$ and $\mat P_s$ are shifting matrices that can shift the order of coefficients in the vectors and matrices. They consist of zero and identity matrices and have a size of $3m\times 3m$ and $3p\times 3p$, respectively. $m$ is the length of sparse code for each color channel, $p$ is the size of HR patches. $\mat D_l \in \mathbb{R}^{3q\times 3m}$ and $\mat D_h \in \mathbb{R}^{3p\times 3m}$ are dictionaries that contain color dictionaries in their block diagonals and
$q$ is length of LR features (patches). We also define and simplify $\mat D_{hs}$:
\bea
    \mat D_{hs} &=&
    \begin{bmatrix}
    \mat D_{h_b}^T \mat S_b^T \mat S_r \mat D_{h_r}   &                          \mat 0                              &                       \mat 0  \\
    \mat 0                                            & \mat D_{h_r}^T \mat S_r^T \mat S_g \mat D_{h_g}              &                       \mat 0   \\
    \mat 0                                            &                         \mat 0                               & \mat D_{h_g}^T \mat S_g^T \mat S_b \mat D_{h_b}    \\
    \end{bmatrix}  \nonumber\\
    &=& \mat P \mat D_h^T \mat S^T \mat P_s^T \mat S \mat D_h \label{Eq:Dhs}
\eea
Finally, the cost function in \eqref{Eq:Cost2} can be written as follows:
\bea
    L_1 &=& \frac{1}{2} \| \vect y_l - \mat D_l \vect x\|_2^2 + \lambda \|\vect x \|_1 \nonumber\\
    &&+ 2 \tau \vect x^T \mat D_h^T \mat S^T \mat S \mat D_h \vect x - 2 \tau \vect x^T \mat P^T \mat D_{hs} \vect x. \label{Eq:Cost3}\\
    &=& \vect x^T [ \frac{1}{2} \mat D_l^T \mat D_l + 2 \tau \mat D_h^T \mat S^T \mat S \mat D_h - 2\tau \mat P^T \mat D_{hs}] \vect x \nonumber\\
    && -\vect y_l^T \mat D_l \vect x + \frac{1}{2} \vect y_l^T \vect y_l + \lambda \|\vect x \|_1
\eea
Substituting \eqref{Eq:Dhs} in the above we have:
\bea
    \vect x^\ast &=~~ \arg\displaystyle\min_{\vect x}& \vect x^T \overbrace{[ \frac{1}{2} \mat D_l^T \mat D_l + 2 \tau \mat D_h^T \mat S^T (\mat I - \mat P_s^T)\mat S \mat D_h ]}^{\mat D} \vect x \nonumber\\
    && -\vect y_l^T \mat D_l \vect x + \frac{1}{2} \vect y_l^T \vect y_l + \lambda \|\vect x \|_1 \\
    &=~~ \arg\displaystyle\min_{\vect x} &\vect x^T \mat D \vect x - \vect y_l^T \mat D_l \vect x \ + \lambda\|\vect x\|_1 \label{Eq:FinalOptProb}
\eea
 The re-written cost function in \eqref{Eq:FinalOptProb}, which is now in a more familiar form, is a convex sparsity constrained optimization and consequently numerical algorithms such as FISTA \cite{Beck:IterativeShrinkageThresholdFISTA_ImagScienSIAM2009, Wright:SpaRSA_TSP2009, Minaee:Segmen_Asilomar2015} can be applied to solve it. Note that matrix $\mat D$ captures cross channel constraints using its off-diagonal blocks.


\subsection{Color adaptive patch processing}
In the previous subsection we presented our color image super resolution framework by exploiting color edge similarities across color channels. However, we should emphasize that not all patches in an image have the same amount of color information and edge similarities. Therefore, any single patch should be treated differently in terms of color constraints. The regularizer parameter $\tau$ can control the emphasis on color edge similarities. Next, we explain our approach to automatically determine $\tau$ in an image/patch adaptive manner.

\begin{figure}
  \centering
  \includegraphics[width=0.8\columnwidth]{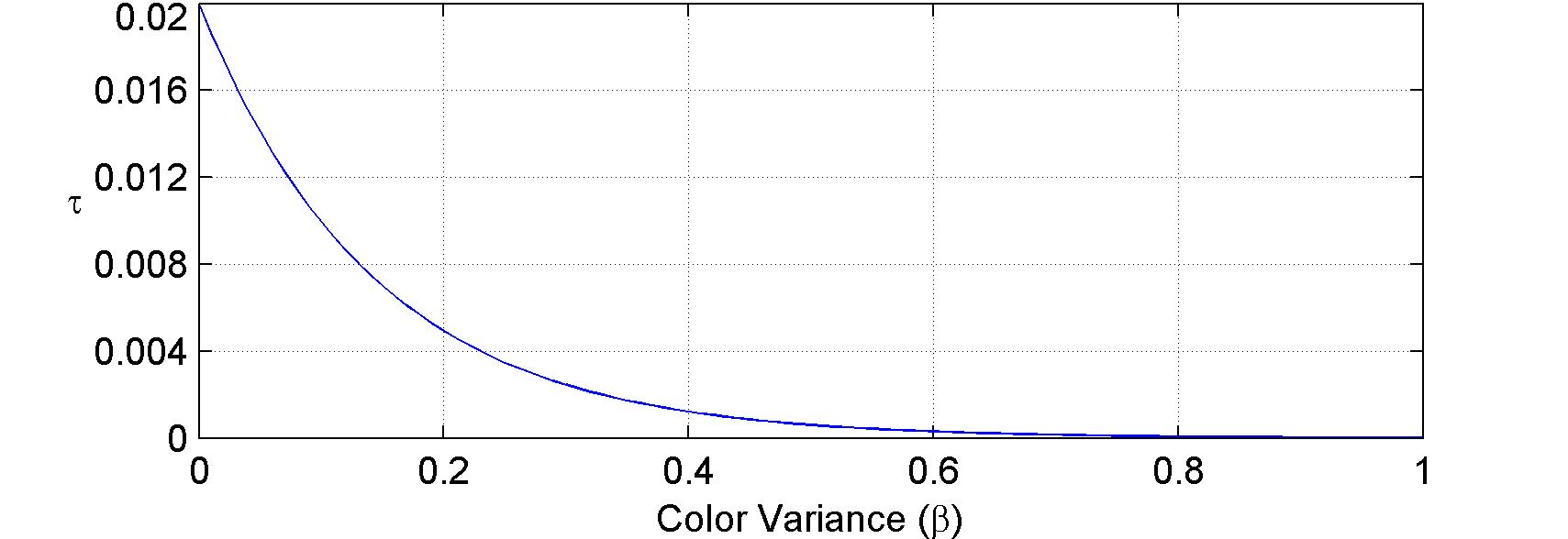}
  \caption{Relationship between color variance $\beta$ and regularizer parameter  $\tau$.}\label{Fig:AdaptivePatch}
\end{figure}

We use the following color variance measure to quantify the color information in each patch:
\bea
    \beta  = \frac{1}{2s} \Big( \frac{\|\mat H_1 \vect y_{Cb} \| + \|\mat H_1 \vect y_{Cr} \|}{\|\mat H_1 \vect y_{Y} \|}  +  \frac{\|\mat H_2 \vect y_{Cb} \| + \|\mat H_2 \vect y_{Cr} \|}{\|\mat H_2 \vect y_{Y} \|} \Big)
\eea
where $s$ is normalization parameter, $\mat H_1$ and $\mat H_2$ are high-pass Scharr operators and $y_Y, y_{Cb}$ and $y_{Cr}$ are Y, Cb and CR channel bands in YCbCr color space. We tested over a large number of image patches and determined a mapping from the $\beta$ values to actual regularizer values ($\tau$) in the optimization framework. This mapping is illustrated in Fig. \ref{Fig:AdaptivePatch}.

\section{Joint learning of Color Dictionaries}
\label{Sec:ColorDL}

Correlation between color channels can be even better captured if the individual color channel dictionaries are also designed to facilitate the same. In order to learn such dictionaries, we propose a new cost function which involves {\em joint learning} of color channel dictionaries.

Given a set of $N$ sampled training image patch pairs $\{\mat Y_h, \mat Y_l\}$, where $\mat Y_h = \{ \vect y_h^1, \vect y_h^2, ..., \vect y_h^N\}$ is the set of high resolution patches sampled from training images and $\mat Y_l = \{ \vect y_l^1, \vect y_l^2, ..., \vect y_l^N\}$ is the set of corresponding low resolution patches or extracted features, we aim to learn dictionaries with aforementioned characteristics. One essential requirement of course is that the sparse representation of low resolution patches and corresponding high resolution patches be the same. At the same time, the high resolution dictionary, which is responsible for reconstructing HR patches, should be designed to capture RGB edge correlations in the super-resolved images. Individually, sparse coding problems in low resolution and high resolution settings may be written as:
\bea
    \mat D_l  &=~~ \arg\displaystyle\min_{\mat D_l,\{\vect x^i\}} & \frac{1}{N} \sum_{i=1}^{N} \frac{1}{2}\| \vect y_l^i - \mat D_l \vect x^i \|_2^2 + \lambda \|\vect x^i\|_1  \nonumber \\
    & \text{st.}& \|\mat D_l(:,k) \|_2^2 \le 1, ~~k=1,2,...,K, \label{Eq:LRcost}
\eea
\bea
    \mat D_h  &=~~ \arg\displaystyle \min_{\mat D_h,\{\vect x^i\}} & \frac{1}{N} \sum_{i=1}^{N}  \frac{1}{2}\| \vect y_h^i - \mat D_h \vect x^i \|_2^2 + \lambda\|\vect x^i\|_1  \nonumber \\
    &&     +\tau \Big[ \|\mat S_r \mat D_{h_r} \vect x_r^i - \mat S_g \mat D_{h_g} \vect x_g^i \|_2^2 \nonumber\\
    &&     ~~+\|\mat S_g \mat D_{h_g} \vect x_g^i - \mat S_b \mat D_{h_b} \vect x_b^i\|_2^2 \nonumber\\
    &&     ~~+\|\mat S_b \mat D_{h_b} \vect x_b^i-\mat S_r \mat D_{h_r} \vect x_r^i\|_2^2 \Big] \nonumber \\
    & \text{st.}& \|\mat D_h(:,k) \|_2^2 \le 1, ~~k=1,2,...,K. \label{Eq:HRcost}
\eea
The additional terms in \eqref{Eq:HRcost} incorporate the edge information across color channels as in \eqref{Eq:MainOptProb}. Note that there is an implicit constraint on $\mat D_l$ and $\mat D_h$ that they both are block diagonal matrices as defined in \eqref{Eq:DicsDef}.  Considering the requirement that the sparse codes are the same for LR and HR framework, we can obtain the following optimization problem which simultaneously optimizes the LR and HR dictionaries:
\bea
     &\arg\displaystyle \min_{\mat D_h, \mat D_l,\{\vect x^i\}} & \frac{1}{N} \sum_{i=1}^{N}  \frac{\gamma}{2}\| \vect y_l^i - \mat D_l \vect x^i \|_2^2 + \frac{1-\gamma}{2}\| \vect y_h^i - \mat D_h \vect x^i \|_2^2 \nonumber \\
    &&     +\tau \Big[ \|\mat S_r \mat D_{h_r} \vect x_r^i - \mat S_g \mat D_{h_g} \vect x_g^i \|_2^2 \nonumber\\
    &&     ~~+\|\mat S_g \mat D_{h_g} \vect x_g^i - \mat S_b \mat D_{h_b} \vect x_b^i\|_2^2 \nonumber\\
    &&     ~~+\|\mat S_b \mat D_{h_b} \vect x_b^i-\mat S_r \mat D_{h_r} \vect x_r^i\|_2^2 \Big] + \lambda \|\vect x^i\|_1 \nonumber \\
    & \text{st.}& \|\mat D_h(:,k) \|_2^2 \le 1, ~\|\mat D_l(:,k) \|_2^2 \le 1,~~ k=1,2,...,K \nonumber\\ \label{Eq:DicCost}
\eea
where $\gamma$ balances the reconstruction error in low resolution and high resolution settings. Using simplifications similar to \eqref{Eq:Cost3}, this cost function can be re-written as follows:
\bea
     L_2&=& \frac{1}{N} \sum_{i=1}^{N}  \frac{\gamma}{2}\| \vect y_l^i - \mat D_l \vect x^i \|_2^2 + \frac{1-\gamma}{2}\| \vect y_h^i - \mat D_h \vect x^i \|_2^2  + \lambda \|\vect x^i\|_1 \nonumber \\
        && +2\tau   \vect x^{i^T} \mat D_h^T \mat S^T (\mat I - \mat P_s^T) \mat S \mat D_h \vect x^{i^T}  \label{Eq:DicCost2} \\
        &=&     \frac{\gamma}{2N}\| \mat Y_l - \mat D_l \mat X \|_F^2 + \frac{1-\gamma}{2N}\| \mat Y_h - \mat D_h \mat X \|_F^2  + \frac{\lambda}{N} \|\mat X\|_1 \nonumber \\
        && +\frac{2\tau}{N} ~\Tr \Big( \mat X^T \mat D_h^T \mat S^T (\mat I - \mat P_s^T) \mat S \mat D_h \mat X^T \Big).~ \label{Eq:MainDicCost}
\eea
where $\mat X = [\vect x^1~ \vect x^2~...~ \vect x^N] \in \mathbb{R}^{3m\times N}$. The first and second terms in \eqref{Eq:MainDicCost} are respectively responsible for small reconstruction error in low resolution and high resolution training data. The third term enforces sparsity and the last one encourages edge similarity via the learned dictionaries. We propose to minimize this cost function by alternatively optimizing over $\mat X, \mat D_l$  and $\mat D_h$ individually, while keeping the others fixed.

With $\mat D_l$ and $\mat D_h$ being fixed, we optimize \eqref{Eq:MainDicCost} over sparse code matrix $\mat X$. Interestingly because of the Trace operator and Frobenius norm, columns of $\mat X$  can be obtained independently. For each column of $\mat X$ ($i=1...N$) we can simplify the problem:
\bea
    \vect x^i &= ~~\arg\displaystyle\min_{\vect x}& \frac{\gamma}{2}\| \vect y_l^i -\mat D_l \vect x \|_F^2 + \frac{1-\gamma}{2}\| \vect y_h^i - \mat D_h \vect x \|_F^2  + \lambda \|\vect x\|_1 \nonumber\\
     &&     +2\tau   \vect x^{T} \mat D_h^T \mat S^T (\mat I - \mat P_s^T) \mat S \mat D_h \vect x^{T}  \nonumber\\
      &=~~ \arg\displaystyle\min_{\vect x}& \vect x^T [ \frac{\gamma}{2} \mat D_l^T \mat D_l +\frac{1-\gamma}{2} \mat D_h^T \mat D_h \nonumber\\
      &&~~~~~~~+ 2 \tau \mat D_h^T \mat S^T (\mat I - \mat P_s^T)\mat S \mat D_h ]  \vect x \nonumber\\
    && -\big( \gamma \vect y_l^{i^T} \mat D_l + (1-\gamma)\vect y_h^{i^T} \mat D_h \big)\vect x + \lambda \|\vect x \|_1  \nonumber\\
    &=~~ \arg\displaystyle\min_{\vect x} &\vect x^T \mat A \vect x - \vect b^T  \vect x \ + \lambda\|\vect x\|_1 \label{Eq:findx}
\eea
where $\mat A =  \frac{\gamma}{2} \mat D_l^T \mat D_l +\frac{1-\gamma}{2} \mat D_h^T \mat D_h + 2 \tau \mat D_h^T \mat S^T (\mat I - \mat P_s^T)\mat S \mat D_h $ and\\
 $\vect b^{i^T} = \gamma \vect y_l^{i^T} \mat D_l + (1-\gamma)\vect y_h^{i^T} \mat D_h $. The optimization in \eqref{Eq:findx} can be solved using FISTA \cite{Beck:IterativeShrinkageThresholdFISTA_ImagScienSIAM2009}.

The next step is to find the low resolution dictionary $\mat D_l$. By fixing $\mat X$ and $\mat D_h$, the cost function reduces to:
\bea
    \mat D_l &= ~~\arg\displaystyle\min_{\mat D_l}& \|\mat Y_l - \mat D_l \mat X\|_F^2 \nonumber \\
    &\text{s.t.}&   \|\mat D_l(:,k) \|_2^2 \le 1,~~ k=1,2,...,K \nonumber\\
    && \mat D_l \text{ is block diagonal as in \eqref{Eq:DicsDef}}.
\eea
Since $\mat D_l$ is block diagonal and there is no explicit cross channel constraint for the low resolution dictionary, the above optimization can be split into three separate dictionary learning procedures as follows where $c\in \{r,g,b\}$.
\bea
    \mat D_{l_c} &= ~~\arg\displaystyle\min_{\mat D_{l_c}}& \|\mat Y_{l_c} - \mat D_{l_c} \mat X_c\|_F^2 \nonumber \\
    &\text{s.t.}&   \|\mat D_{l_c}(:,k) \|_2^2 \le 1,~~ k=1,2,...,K \label{Eq:findDl}
\eea
which $\mat X_c = [\vect x_c^1~ \vect x_c^2~...~ \vect x_c^N] \in \mathbb{R}^{m\times N}$, $\mat Y_{l_c} = [\vect y_c^1~ \vect y_c^2~...~ \vect y_c^N]\in \mathbb{R}^{p\times N}$ and $c$ takes the subscripts from $\{r,g,b\}$ indicating a specific color channel.
Each of the above dictionaries are learned by the dictionary learning  method in \cite{Mairal:ODL_ICML2009}.

Finally, for finding $\mat D_h$, when $\mat X$ and $\mat D_l$ are fixed, we have:
\bea
     \mat D_h &= ~~ \arg\displaystyle\min_{\mat D_{h}}& \frac{1}{N} \sum_{i=1}^{N}   \frac{1-\gamma}{2}\| \vect y_h^i - \mat D_h \vect x^i \|_2^2  \nonumber\\
     &&~~+ 2\tau   \vect x^{i^T} \mat D_h^T \mat S^T (\mat I - \mat P_s^T) \mat S \mat D_h \vect x^{i^T} \nonumber\\
        &\text{s.t}&  \|\mat D_{h}(:,k) \|_2^2 \le 1,~~ k=1,2,...,K.    \nonumber\\
        && \mat D_h \text{ is block diagonal as in \eqref{Eq:DicsDef}}\label{Eq:findDh}
\eea
We develop a solution for (\ref{Eq:findDh}) using the Alternative Direction Method of Multipliers (ADMM) \cite{Boyd:ADMM_MachineLearn2011}.
We first define the function $g(\mat D_h, \mat Z)$ as follows which is essentially the same cost function with the multiplication by $\mat D_h$ in the final term of (\ref{Eq:findDh}) substituted by a slack matrix $\mat Z$:
\bea
    g(\mat D_h, \mat Z) = \frac{1}{N} \sum_{i=1}^{N}   \frac{1-\gamma}{2}\| \vect y_h^i - \mat D_h \vect x^i \|_2^2 + 2\tau   \vect x^{i^T} \mat D_h^T \mat S^T (\mat I - \mat P_s^T) \mat S \mat Z \vect x^{i^T}\nonumber
\eea
Then, solving the following optimization problem, which is a bi-convex problem, is equivalent to solving \eqref{Eq:findDh}.
\bea
    \mat D_h &= ~~ \arg\displaystyle\min_{\mat D_{h},\mat Z}& g(\mat D_h, \mat Z) \nonumber\\
    &\text{s.t}& \mat D_h - \mat Z = \mat 0,\nonumber\\
    &&              \|\mat D_{h}(:,k) \|_2^2 \le 1,~~ k=1,2,...,K.\nonumber\\
    && \mat D_h \text{ is block diagonal as in \eqref{Eq:DicsDef}}. \label{Eq:EquivalenDicOpt}
\eea
The following is a summary of iterative solution to \eqref{Eq:EquivalenDicOpt} using ADMM until a convergence is achieved where $t$ is the iteration index of ADMM procedure:

    \bea
        1)~~~\mat D_h^{t+1} &=~~ \arg\displaystyle\min_{\mat D_{h}}& \Big( \frac{1}{N} \sum_{i=1}^{N}   \frac{1-\gamma}{2}\| \vect y_h^i - \mat D_h \vect x^i \|_2^2  \nonumber\\
         &&~~+ 2\tau   \vect x^{i^T} \mat D_h^T \mat S^T (\mat I - \mat P_s^T) \mat S \mat Z^t \vect x^{i^T} \Big) \nonumber\\
         &&~~+ \frac{\rho}{2} \| \mat D_h -  \mat Z^t + \mat U^t\|_F^2 \nonumber\\
         &s.t.& \|\mat D_{h}(:,k) \|_2^2 \le 1, ~k=1,...,K.\nonumber\\
         &&     \mat D_h \text{ is block diagonal as \eqref{Eq:DicsDef}.} \label{Eq:ADMMstep1}
    \eea
    \bea
        2)~~~\mat Z^{t+1} &= \arg\displaystyle\min_{\mat Z}& \Big( \frac{2\tau}{N} \sum_{i=1}^{N} \vect x^{i^T} \mat D_h^{t+1^T} \mat S^T (\mat I - \mat P_s^T) \mat S \mat Z^t \vect x^{i^T} \Big) \nonumber\\
         &&~~+ \frac{\rho}{2} \| \mat D_h^{t+1} -  \mat Z + \mat U^t\|_F^2 \label{Eq:ADMMstep2}
    \eea
    \bea
        3)~~~\mat U^{t+1} &=& \mat U^{t} + \mat D_h^{t+1} - \mat Z^{t+1} ~~~~~~~~~~~~~~~~~~~~~~\label{Eq:ADMMstep3}
    \eea
Step 3 of the above ADMM procedure is straight forward. However, Steps 1 and 2 need further analytical simplifications for tractability.

\textbf{Step 1:} The optimization  in this step can be re-written as:
\bea
    \mat D_h^{t+1} &=~~ \arg\displaystyle\min_{\mat D_{h}}& \Tr (\mat D_h \mat F \mat D_h^T) - 2 \Tr (\mat E \mat D_h^T) \nonumber\\
    &\text{s.t.} & \|\mat D_{h}(:,k) \|_2^2 \le 1, ~k=1,...,K. \nonumber\\
    &&\mat D_h \text{ is block diagonal as in \eqref{Eq:DicsDef}}   \label{Eq:DhADMM}
\eea
where
\bea
    \mat F &=& \frac{1-\gamma}{2N} \mat X \mat X^T   + \frac{\rho}{2} \mat I_{3m\times 3m} \\
    \mat E &=& \frac{1-\gamma}{2N} \mat Y_h \mat X^T +\frac{\rho}{2} (\mat Z^k - \mat U^k) -\frac{\tau}{N} \mat S^T (\mat I -\mat P_s^T) \mat S \mat Z^k \mat X \mat X^T. \nonumber\\
\eea
Assuming the following block structure for $\mat E$ and $\mat F$:
\bea
    \mat F =
        {\begin{bmatrix}
        \mat F_{rr}       & \mat F_{12}     & \mat F_{13}      \\
        \mat F_{21}       & \mat F_{gg}     & \mat F_{23}   \\
        \mat F_{31}       & \mat F_{32}     & \mat F_{bb} \\
        \end{bmatrix}},~~
    \mat E =
        {\begin{bmatrix}
        \mat E_{rr}       & \mat E_{12}     & \mat E_{13}      \\
        \mat E_{21}       & \mat E_{gg}     & \mat E_{23}   \\
        \mat E_{31}       & \mat E_{32}     & \mat E_{bb} \\
        \end{bmatrix}}
\eea
and due to the block diagonal structure of $\mat D_h$ as in \eqref{Eq:DicsDef}, we can rewrite each term in \eqref{Eq:DhADMM} in the following form:
\bea
    \Tr (\mat E \mat D_h^T) = \Tr (\mat E_{rr} \mat D_{h_r}^T) +\Tr (\mat E_{gg} \mat D_{h_g}^T) +\Tr (\mat E_{bb} \mat D_{h_b}^T)
\eea
\bea
    \Tr (\mat D_h \mat F \mat D_h^T) = \Tr (\mat D_{h_r} \mat F_{rr} \mat D_{h_r}^T) +\Tr (\mat D_{h_g} \mat F_{gg} \mat D_{h_g}^T) \nonumber\\ ~~~~+\Tr (\mat D_{h_b} \mat F_{bb} \mat D_{h_b}^T)
\eea
Finally the cost function reduces to:
\bea
    & \arg\displaystyle\min_{\mat D_{h_r},\mat D_{h_g},\mat D_{h_b}}& \Tr(\mat D_{h_r} \mat F_{rr} \mat D_{h_r}^T) -2 \Tr (\mat E_{rr} \mat D_{h_r}^T) \nonumber\\
    &&+  \Tr(\mat D_{h_g} \mat F_{gg} \mat D_{h_g}^T) -2 \Tr (\mat E_{gg} \mat D_{h_g}^T) \nonumber\\
    &&+  \Tr(\mat D_{h_b} \mat F_{bb} \mat D_{h_b}^T) -2 \Tr (\mat E_{bb} \mat D_{h_b}^T) \nonumber\\
    &\text{s.t.} & \|\mat D_{h_c}(:,k) \|_2^2 \le 1, ~ c\in \{r,g,b\}.
\eea
which is a separable optimization problem, i.e. it can be solved for $\mat D_{h_r}, \mat D_{h_g}$ and $\mat D_{h_b}$ separately as follows:
 \bea
    & \arg\displaystyle\min_{\mat D_{h_c}}& \Tr(\mat D_{h_c} \mat F_{cc} \mat D_{h_c}^T) -2 \Tr (\mat E_{cc} \mat D_{h_c}^T) \nonumber\\
    &\text{s.t.} & \|\mat D_{h_c}(:,k) \|_2^2 \le 1, ~ k=1,2,...,K_c.
\eea
Each of above subproblems now is solvable using the algorithmic approach in Online Dictionary Learning \cite{Mairal:ODL_ICML2009}.

\textbf{Step 2:} This is an {\em unconstrained} convex optimization problem in terms of $\mat Z$ and we can find the minimum by taking the derivative. The closed form solution for $\mat Z$ is given by:
\bea
    \mat Z^{t+1} = \mat D_h^{t+1} + \mat U^{t+1} - \frac{2\tau}{N\rho}  \mat S^T (\mat I - \mat P_s) \mat S \mat D_h^{t+1} \mat X \mat X^T
\eea

A formal stepwise description of our color dictionary learning algorithm is given in Algorithm \ref{Alg:ColorDL}.

\begin{algorithm}[t]
\caption{Color Dictionary Learning }
\label{Alg:ColorDL}
\begin{algorithmic}
\REQUIRE $\mat Y_l, \mat Y_h, \tau, \lambda, \rho$.\\
\emph{initialize: } $\mat D_h^0, \mat D_l^0$, iteration index $n=1$.
\FOR{$n=1: $ Maxiter}
\STATE(1) Find the sparse code matrix by Solving the convex optimization problem in \eqref{Eq:findx}:
\STATE(2) Solve the LR dictionary learning problem in \eqref{Eq:findDl}
\STATE(3) Solve the HR dictionary learning problem in \eqref{Eq:findDh}:
    \WHILE{stopping criterion not met} 
    \STATE(3-1) Solve for $\mat D_h^{t+1}$ using \eqref{Eq:ADMMstep1}
    \STATE(3-2) Solve for $\mat Z^{t+1}$ using \eqref{Eq:ADMMstep2}
    \STATE(3-3) Solve for $\mat U^{t+1}$ using \eqref{Eq:ADMMstep3}
    \STATE(3-4) Increase inner iteration index $t$.
   \ENDWHILE{ if $\| \mat D_h^{t+1} - \mat D_h^{t}\|_F < tol$ }
\STATE(4) Increase iteration index $n$.
\ENDFOR
\ENSURE $\mat D_h, \mat D_l$.
\end{algorithmic}
\end{algorithm}

\section{Experimental Results}
\label{Sec:Experiments}
Our experiments are performed on the widely used \emph{set 5} and \emph{set 14} images as in \cite{Zeyde:SR_Springer2012}. We compare the proposed Multi-Channel constrained Super Resolution (MCcSR) method with several well-known single image super resolution methods. These include the ScSR \cite{Yang:CoupledDicLearnSR_TIP2012} method because our MCcSR method can be seen as a multi-channel extension of the same. Other methods for which we report results are the Single Image Scale-up using Sparse Representation by Zeyde \emph{et al.} \cite{Zeyde:SR_Springer2012}, Anchored Neighborhood Regression for Fast Example-Based Super-Resolution (ANR) \cite{Timofte:AnchoredARN+_ACCV2014} and Global Regression (GR) \cite{Timofte:AnchoredANR_ICCV2013} methods by Timofte \emph{et al}, Neighbor Embedding with Locally Linear Embedding (NE+LLE) \cite{Chang:NeighborEmbeddingSR_CVPR2004} and Neighbor Embedding with NonNegative Least Squares (NE+NNLS) \cite{Bevilacqua:NENNLS_BMVA2012} that were both adapted to learned dictionaries.

In our experiments, we will magnify the input images by a factor of $2$, $3$ or $4$, which is commonplace in the literature. For the low-resolution images, we
use $5 \times 5$ low-resolution patches with overlap of 4 pixels between adjacent patches and extract features based on method in \cite{YangAndWright:SparseSR_TIP2010}. It is noteworthy to mention that these features are not extracted from the $5 \times 5$ low resolution patches, but rather from bicubic interpolated version of the whole image with the desired magnification factor. Extracted features are then used to find the sparse codes according to \eqref{Eq:FinalOptProb} which involves color information as well. Then, high resolution patches are reconstructed based on the same sparse code using the learned high resolution dictionaries and averaged over the overlapping regions. Dictionaries are obtained by training over $100000$ patch pairs which are preprocessed by cropping out the textured regions and discarding the smooth regions. The number of columns in each learned dictionary is $512$ for most of our experiments and regularization parameter $\lambda$ is picked via cross-validation to be $0.1$.
\begin{figure*}
  \centering
  \includegraphics [width = 1\textwidth]{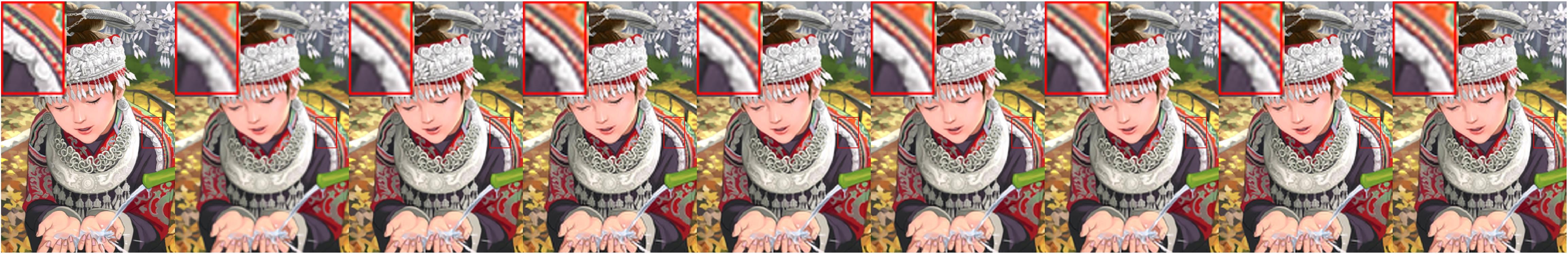}
  \includegraphics [width = 1\textwidth]{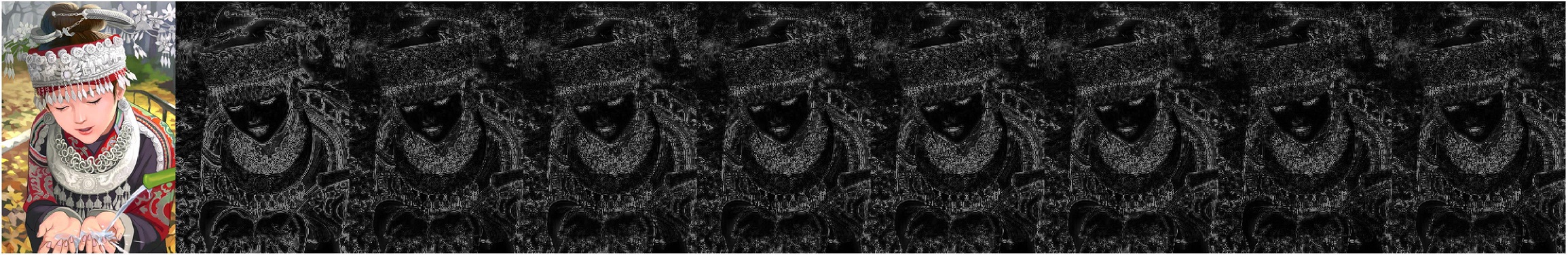}
  \caption{Comparison of different methods for comic image with scaling factor of 2 (Please refer to the electronic version and zoom in for obvious comparison). Numbers in parenthesis are PSNR, SSIM and SCIELAB error measures, respectively. Left to right: Original,
                    Bicubic      (30.46, 0.840, 1.898e4),
                    Zeyde et al. (31.97, 0.887, 1.127e4),
                    GR           (31.70, 0.879, 1.198e4),
                    ANR          (32.09, 0.889, 1.077e4),
                    NENNLS       (31.87, 0.884, 1.159e4),
                    NELLE        (32.03, 0.889, 1.099e4),
                    \tb{MCcSR    (32.23, 0.899, 9.770e3)},
                    ScSR         (32.14, 0.893, 1.014e4). }
    \label{Fig:Comic2x}
\end{figure*}
\begin{figure*}
  \centering
  \includegraphics [width = 1\textwidth]{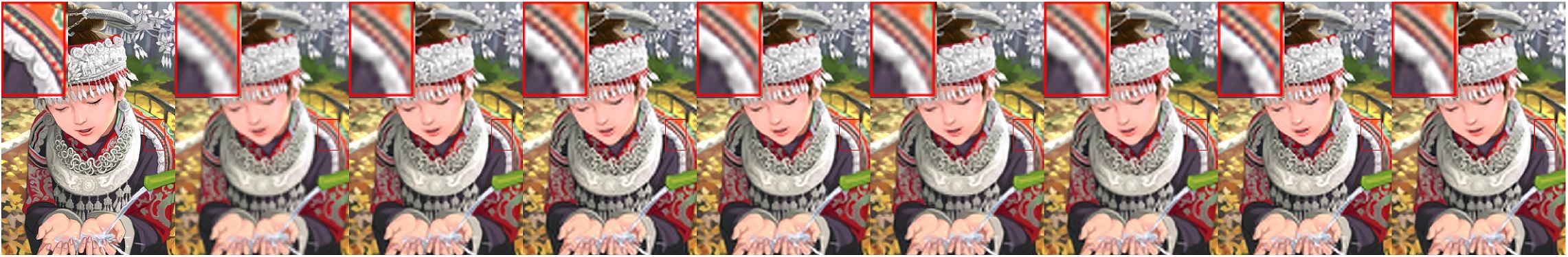}
  \includegraphics [width = 1\textwidth]{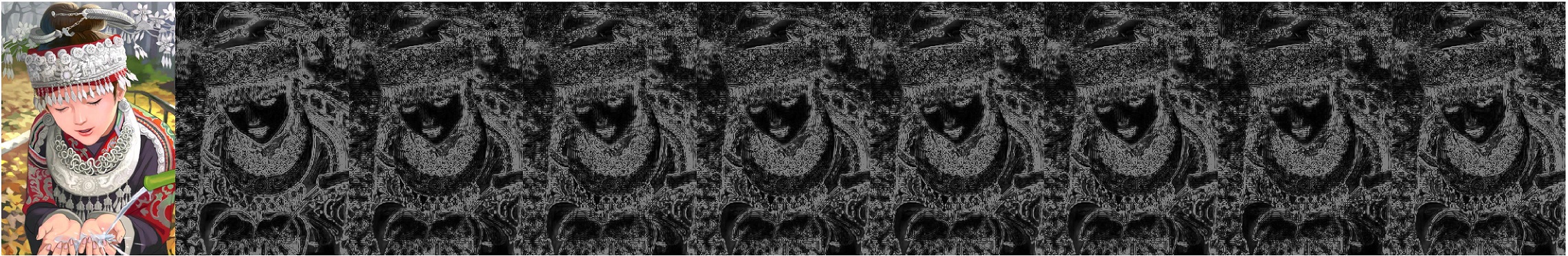}
       \caption{ Super-resolution results for scaling factor 3 and quantitative measures. Left to right:
    {               Original,
                    Bicubic      (27.51, 0.685, 3.423e4),
                    Zeyde et al. (28.28, 0.737, 2.896e4),
                    GR           (28.15, 0.729, 3.008e4),
                    ANR          (28.36, 0.742, 2.865e4),
                    NENNLS       (28.17, 0.730, 2.961e4),
                    NELLE        (28.30, 0.738, 2.905e4),
                    \tb{MCcSR    (28.51, 0.758, 2.709e4)},
                    ScSR         (28.31, 0.740, 2.860e4) . }}
        \label{Fig:Comic3x}
\end{figure*}
\begin{figure*}
  \centering
  \includegraphics [width = 1\textwidth]{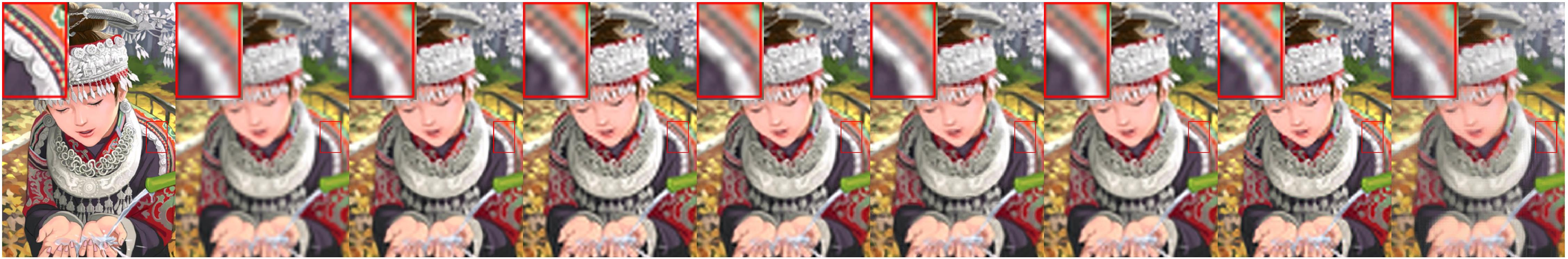}
  \includegraphics [width = 1\textwidth]{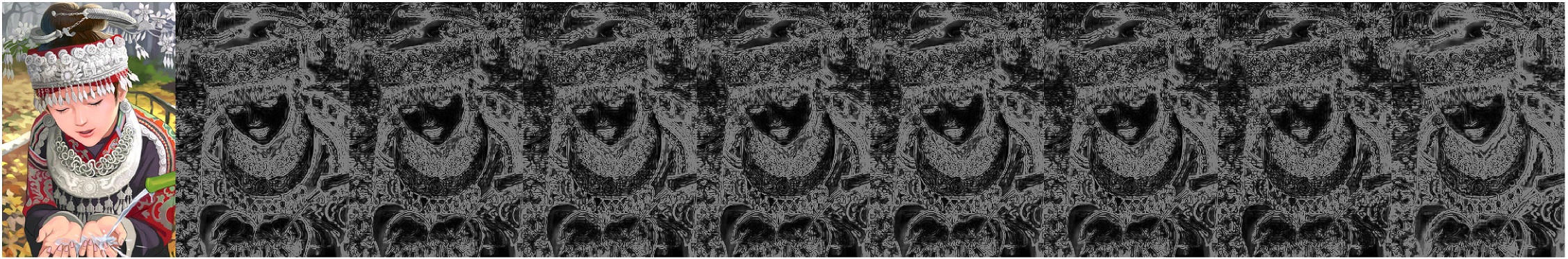}
     \caption{  Super-resolution results for scaling factor 4 and quantitative measures. Left to right:
    {               Original,
                    Bicubic      (26.05, 0.566, 4.369e4),
                    Zeyde et al. (26.61, 0.615, 3.923e4),
                    GR           (26.51, 0.607, 4.045e4),
                    ANR          (26.63, 0.618, 3.928e4),
                    NENNLS       (26.50, 0.606, 3.984e4),
                    NELLE        (26.57, 0.614, 3.967e4),
                    \tb{MCcSR    (26.74, 0.632, 3.818e4)},
                    ScSR         (26.35, 0.608, 4.002e4) . }}
      \label{Fig:Comic4x}
\end{figure*}

\begin{figure*}
  \centering
  \includegraphics [width = 1\textwidth]{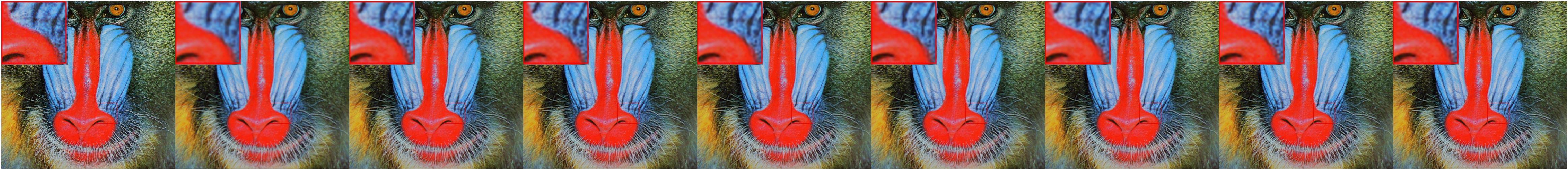}
  \includegraphics [width = 1\textwidth]{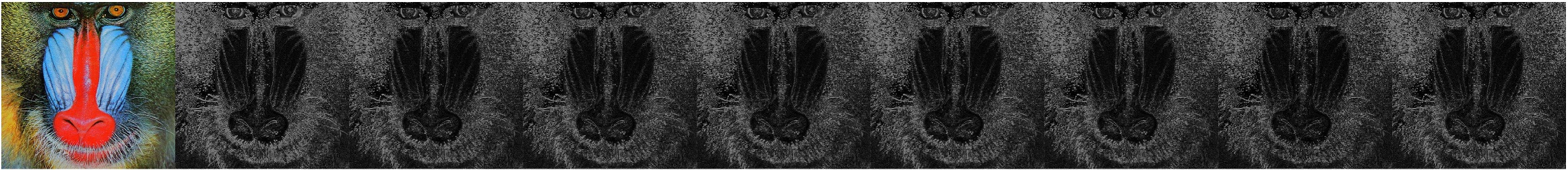}
        \caption{  Comparison of different methods for baboon image with scaling factor of 2. Numbers in parenthesis are PSNR, SSIM and SCIELAB error measures, respectively. Left to right:
    {               Original,
                    Bicubic      (28.19, 0.635, 7.856e4),
                    Zeyde et al. (28.62, 0.683, 6.570e4),
                    GR           (28.63, 0.690, 6.388e4),
                    ANR          (28.67, 0.689, 3.287e4),
                    NENNLS       (28.58, 0.680, 6.585e4),
                    NELLE        (28.66, 0.688, 6.421e4),
                    \tb{MCcSR    (28.78, 0.705, 5.799e4)},
                    ScSR         (28.69, 0.692, 6.296e4) . }}
                    \label{Fig:Baboon2x}
\end{figure*}
\begin{figure*}
  \centering
  \includegraphics [width = 1\textwidth]{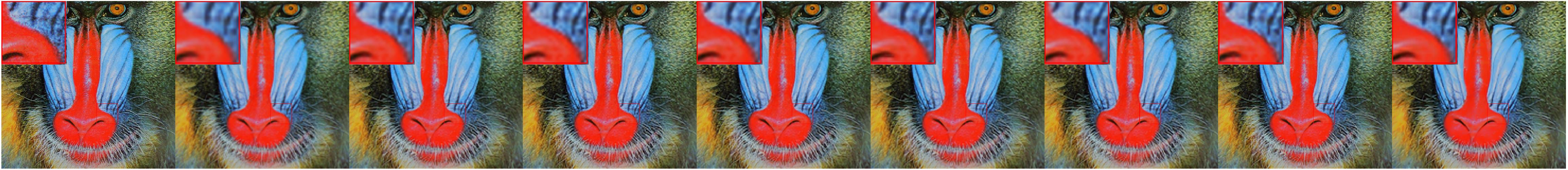}
  \includegraphics [width = 1\textwidth]{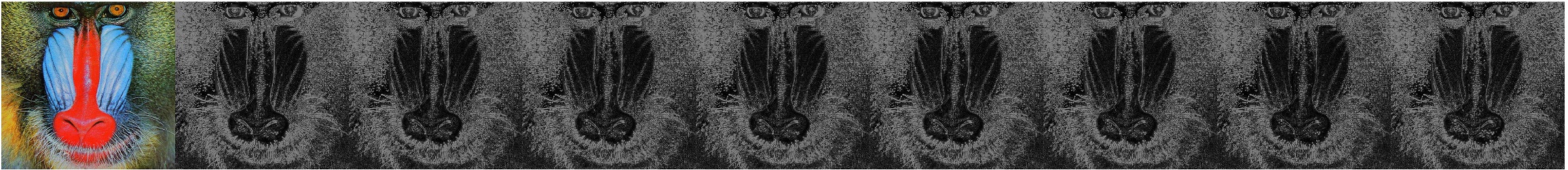}
       \caption{  Super-resolution results for scaling factor 3 and quantitative measures. Left to right:
    {               Original,
                    Bicubic      (26.71, 0.480, 1.078e5),
                    Zeyde et al. (26.94, 0.520, 1.008e5),
                    GR           (26.95, 0.529, 1.000e5),
                    ANR          (26.97, 0.527, 9.962e4),
                    NENNLS       (26.92, 0.518, 1.010e5),
                    NELLE        (26.97, 0.526, 9.998e4),
                    \tb{MCcSR    (27.11, 0.549, 9.574e4)},
                    ScSR         (26.95, 0.524, 1.018e5) . }}
                    \label{Fig:Baboon3x}
\end{figure*}
\begin{figure*}
  \centering
  \includegraphics [width = 1\textwidth]{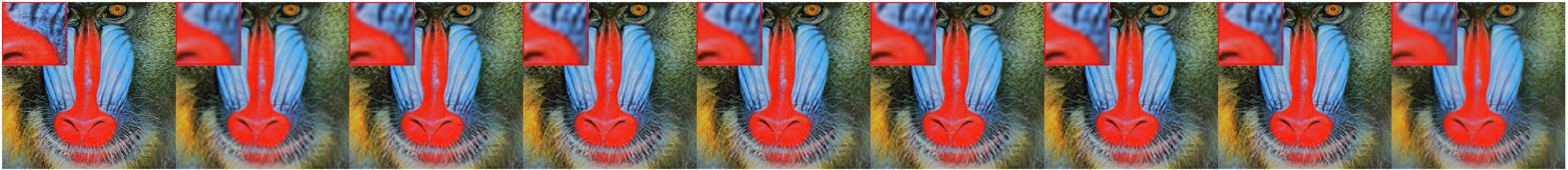}
  \includegraphics [width = 1\textwidth]{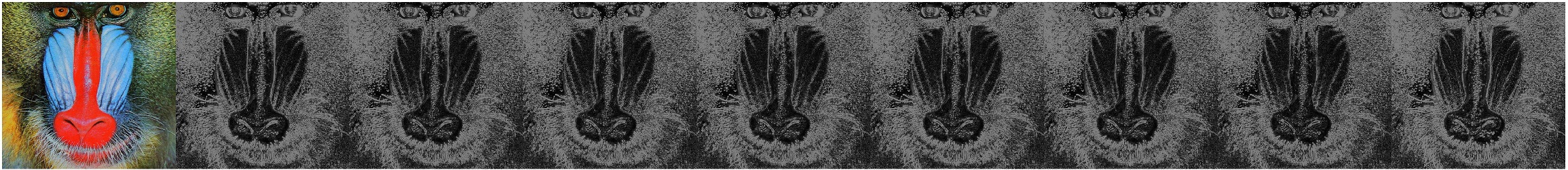}
     \caption{  Super-resolution results for scaling factor 4 and quantitative measures. Left to right:
     {              Original,
                    Bicubic      (26.00, 0.390, 1.237e5),
                    Zeyde et al. (26.17, 0.420, 1.186e5),
                    GR           (26.17, 0.428, 1.183e5),
                    ANR          (26.19, 0.426, 1.180e5),
                    NENNLS       (26.15, 0.419, 1.190e5),
                    NELLE        (26.18, 0.425, 1.183e5),
                    \tb{MCcSR    (26.25, 0.446, 1.136e5)},
                    ScSR         (26.11, 0.415, 1.185e5) . }}
                    \label{Fig:Baboon4x}
\end{figure*}
\begin{figure*}
  \begin{subfigure}[t]{0.33\textwidth}
      \centering
      \includegraphics[width=\textwidth]{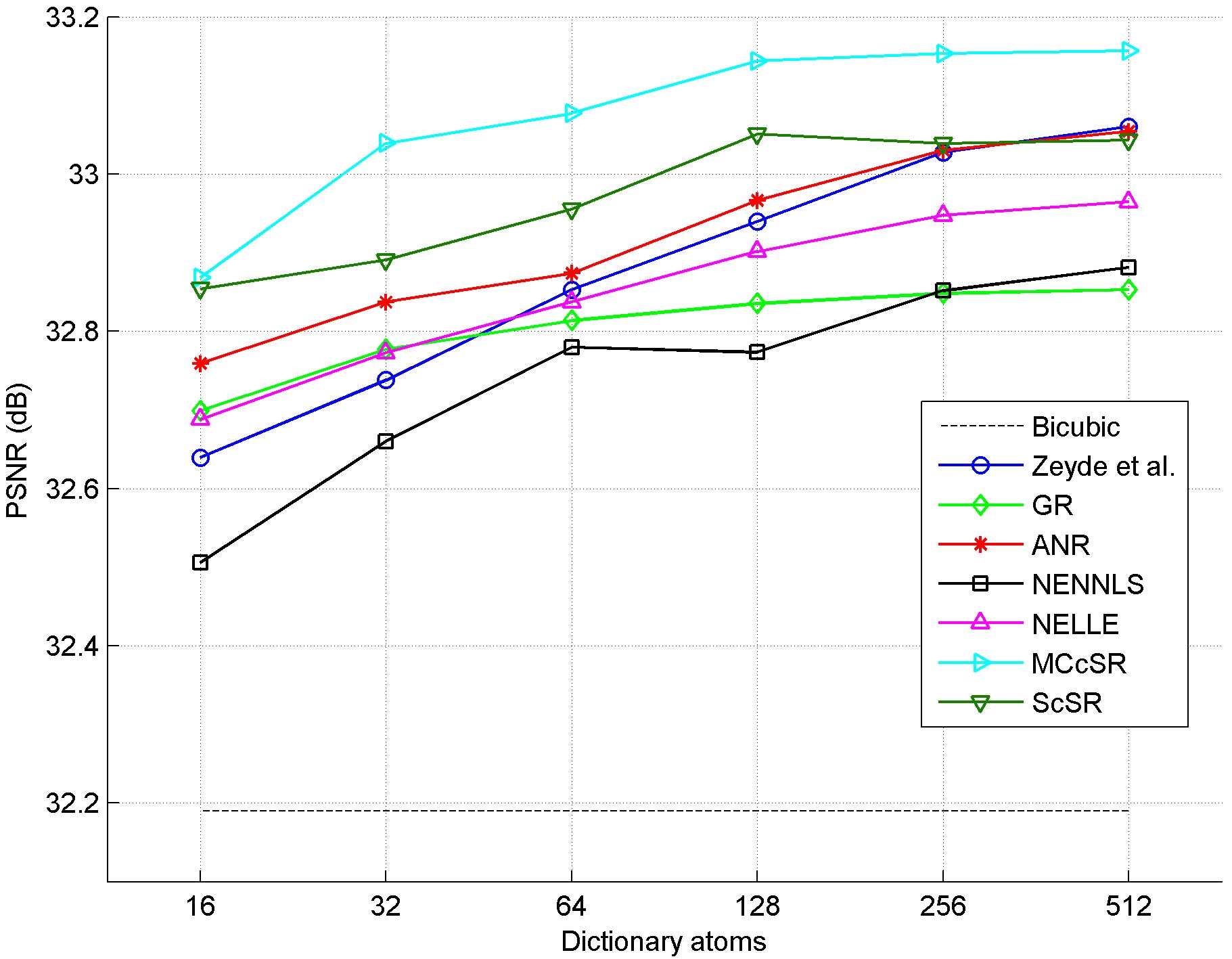}
  \end{subfigure}
  \begin{subfigure}[t]{0.33\textwidth}
      \centering
      \includegraphics[width=\textwidth]{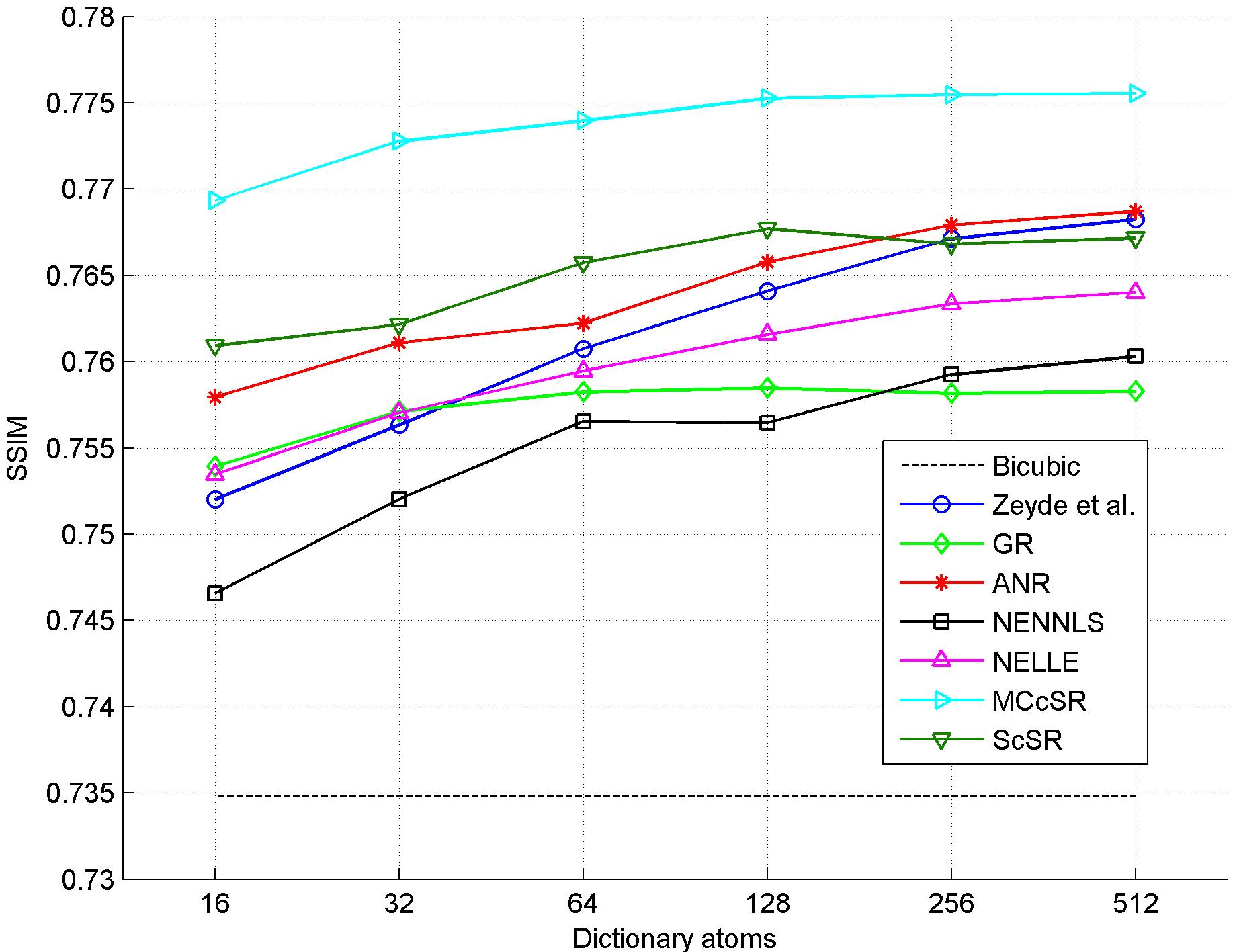}
  \end{subfigure}
  \begin{subfigure}[t]{0.33\textwidth}
      \centering
      \includegraphics[width=\textwidth]{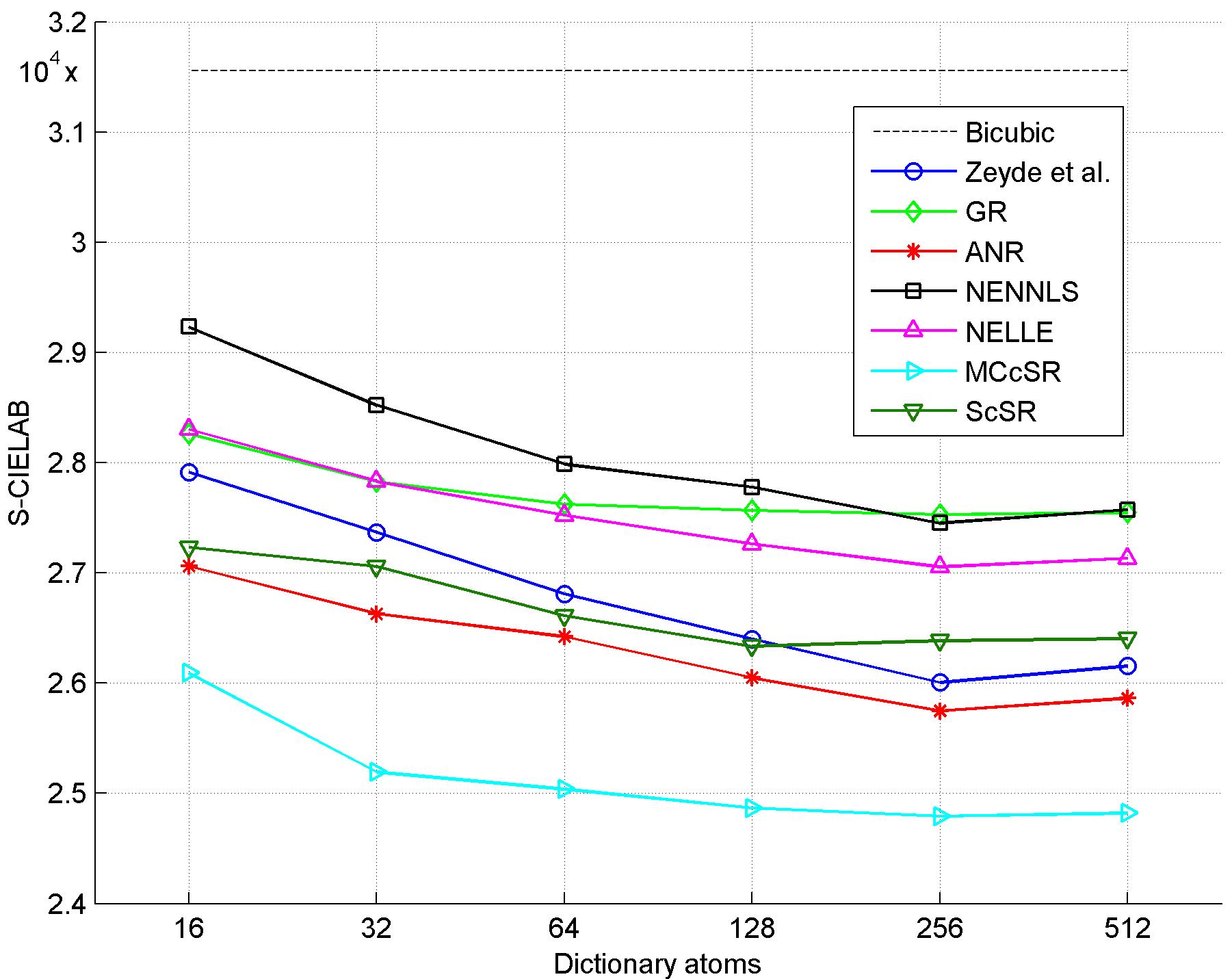}
  \end{subfigure}
  \caption{Effect of dictionary size on PSNR, SSIM and S-CIELAB error of SR methods with a scaling factor of $3$.}\label{Fig:DictSizePSNR}
 \end{figure*}
\begin{table*}
\centering
\begingroup
      \caption{PSNR results of different methods for various images with scaling factor of $3$.}
\label{Tab:imagesPSNR}
      \begin{tabular}{ |l||c|c|c|c|c|c|c|c| }
\hline	
\multirow{2}{*}{Images}  & \multicolumn{8}{c|}{PSNR (dB)} 	\\															 		
      \cline{2-9}
                         &   Bicub  &	Zeyde   &	GR	    &   ANR	    &   NENNLS	&   NELLE   &	MCcSR   &	ScSR        \\ \hline
baby		             &   38.42  &	39.51   &	39.38   &\tb{39.56} &	39.22	&   39.49   &	39.51   &	39.40       \\ \hline
butterfly	             &   28.73  &	30.60   &	29.73   &	30.57   &	30.29	&   30.42   &	30.59   &\tb{30.64}     \\ \hline
bird		             &   36.37  &	37.90   &	37.44   &	37.92   &	37.68	&   37.90   &\tb{38.02} &	37.59       \\ \hline
face		             &   35.96  &	36.44   &	36.40   &\tb{36.50} &	36.39	&   36.47   &	36.48   &	36.37       \\ \hline
foreman	                 &   35.76  &	37.67   &	36.84   &	37.71   &	37.37	&   37.69   &\tb{37.74} &	37.64       \\ \hline
coastguard	             &   31.31  &	31.91   &	31.78   &	31.84   &	31.77	&   31.83   &\tb{31.95} &	31.83       \\ \hline
flowers	                 &   30.92  &	31.84   &	31.62   &	31.88   &	31.68	&   31.80   &\tb{32.07} &	31.87       \\ \hline
head		             &   36.02  &	36.47   &	36.42   &\tb{36.52} &	36.40	&   36.50   &	36.51   &	36.42       \\ \hline
lenna		             &   35.26  &	36.23   &	35.99   &	36.29   &	36.11	&   36.24   &\tb{36.33} &	36.14       \\ \hline
man		                 &   31.78  &	32.68   &	32.44   &	32.71   &	32.50	&   32.65   &\tb{32.75} &	32.68       \\ \hline
pepper		             &   35.25  &	36.27   &	35.77   &	36.13   &	35.99	&   36.12   &\tb{36.30} &	36.20       \\ \hline \hline
average                  &   33.08  &	34.06   &	33.76   &	34.07   &	33.88	&   34.03   &\tb{34.14} &	34.00       \\ \hline
      \end{tabular}
\endgroup
    \end{table*}
\begin{table*}
\centering
\begingroup
      \caption{SSIM results of different methods for various images with scaling factor of $3$.}
\label{Tab:imagesSSIM}
      \begin{tabular}{ |l||c|c|c|c|c|c|c|c| }
\hline	
\multirow{2}{*}{Images}  & \multicolumn{8}{c|}{SSIM} 	\\															 		
      \cline{2-9}
                         &   Bicub	&   Zeyde	&	GR	    &	ANR	    &	NENNLS	&	NELLE	&	MCcSR	&	ScSR        \\ \hline
baby		             &   0.88	&   0.90	&	0.90	&	0.90	&	0.89	&	0.90	&	0.90	&	0.89        \\ \hline
butterfly	             &   0.79	&   0.85	&	0.80	&	0.84	&	0.84	&	0.84	&	0.85	&	0.85        \\ \hline
bird		             &   0.90	&   0.92	&	0.91	&	0.92	&	0.92	&	0.92	&	0.93	&	0.91        \\ \hline
face		             &   0.72	&   0.74	&	0.74	&	0.74	&	0.74	&	0.74	&	0.75	&	0.74        \\ \hline
foreman	                 &   0.89	&   0.91	&	0.90	&	0.91	&	0.90	&	0.91	&	0.91	&	0.90        \\ \hline
coastguard	             &   0.57	&   0.62	&	0.63	&	0.62	&	0.61	&	0.62	&	0.63	&	0.62        \\ \hline
flowers	                 &   0.77	&   0.80	&	0.79	&	0.80	&	0.79	&	0.80	&	0.81	&	0.80        \\ \hline
head		             &   0.72	&   0.74	&	0.74	&	0.75	&	0.74	&	0.74	&	0.75	&	0.74        \\ \hline
lenna		             &   0.78	&   0.80	&	0.80	&	0.80	&	0.80	&	0.80	&	0.81	&	0.80        \\ \hline
man		                 &   0.72	&   0.76	&	0.76	&	0.77	&	0.76	&	0.76	&	0.76	&	0.76        \\ \hline
pepper		             &   0.78	&   0.80	&	0.79	&	0.80	&	0.79	&	0.79	&	0.80	&	0.79        \\ \hline \hline
average                  &   0.745	&   0.776	&   0.769	&   0.778	&   0.771	&   0.775	&\tb{0.785}	&   0.774       \\ \hline
      \end{tabular}
\endgroup
    \end{table*}
\begin{table*}
\centering
\begingroup
      \caption{S-CIELAB error results of different methods for various images with scaling factor of $3$.}
\label{Tab:imagesSCIELAB}
      \begin{tabular}{ |l||c|c|c|c|c|c|c|c| }
\hline
\multirow{2}{*}{Images}  & \multicolumn{8}{c|}{S-CIELAB} 	\\	
      \cline{2-9}
                         &   Bicub	    &   Zeyde	    &   GR	        &   ANR	        &   NENNLS	    &   NELLE	    &   MCcSR	    &   ScSR            \\ \hline
baby		             &   2.07E+04	&   1.36E+04	&   1.40E+04	&\tb{1.32E+04}	&   1.47E+04	&   1.34E+04	&   1.34E+04	&   1.50E+04        \\ \hline
butterfly	             &   2.28E+04	&   1.55E+04	&   1.84E+04	&   1.55E+04	&   1.60E+04	&   1.60E+04	&   1.54E+04	&\tb{1.49E+04}      \\ \hline
bird		             &   1.07E+04	&   7.36E+03	&   8.02E+03	&   7.21E+03	&   7.73E+03	&   7.30E+03	&\tb{6.50E+03}	&   7.81E+03        \\ \hline
face		             &   3.79E+03	&   2.71E+03	&   2.73E+03	&   2.57E+03	&   2.73E+03	&   2.61E+03	&\tb{2.47E+03}	&   2.70E+03        \\ \hline
foreman	                 &   8.46E+03	&   3.90E+03	&   4.79E+03	&\tb{3.48E+03}	&   4.01E+03	&   3.62E+03	&   3.72E+03	&   3.89E+03        \\ \hline
coastguard	             &   1.96E+04	&   1.71E+04	&   1.70E+04	&   1.70E+04	&   1.76E+04	&   1.71E+04	&\tb{1.69E+04}	&   1.70E+04        \\ \hline
flowers	                 &   4.47E+04	&   3.75E+04	&   3.89E+04	&   3.69E+04	&   3.84E+04	&   3.74E+04	&\tb{3.29E+04}	&   3.70E+04        \\ \hline
head		             &   3.79E+03	&   2.69E+03	&   2.74E+03	&   2.54E+03	&   2.79E+03	&   2.61E+03	&\tb{2.42E+03}	&   2.65E+03        \\ \hline
lenna		             &   2.44E+04	&   1.74E+04	&   1.85E+04	&   1.67E+04	&   1.79E+04	&   1.69E+04	&\tb{1.58E+04}	&   1.72E+04        \\ \hline
man		                 &   3.80E+04	&   2.91E+04	&   3.03E+04	&\tb{2.84E+04}	&   3.02E+04	&   2.89E+04	&   2.88E+04	&   2.95E+04        \\ \hline
pepper		             &   2.48E+04	&   1.91E+04	&   2.15E+04	&   1.96E+04	&   2.02E+04	&   1.95E+04	&\tb{1.73E+04}	&   1.91E+04        \\ \hline \hline
average                  &   2.79E+04	&   2.27E+04	&   2.36E+04	&   2.24E+04	&   2.33E+04	&   2.26E+04	&\tb{2.14E+04}	&   2.28E+04        \\ \hline
      \end{tabular}
\endgroup
    \end{table*}

%

We perform visual comparisons of obtained super-resolution images and additionally evaluate them quantitatively using image quality metrics. The metrics we use include: 1.) Peak Signal to Noise Ratio (PSNR) while recognizing its limitations \cite{Wang:HowGoodIsMSE_SPM2009}\footnote{Note that since we  work on color images, the PSNR reported is carried out on all the color channels.}, 2.) the widely used Structural Similarity Index (SSIM) \cite{Wang:SSIM_TIP2004} and 3.) a popular color-specific quality measure called S-CIELAB \cite{Zhang:SCIELAB_Elsevier1998} which evaluates color fidelity while taking spatial context into account.

\subsection{Generic SR results}
Fig. \ref{Fig:Comic2x} show SR results for a popular natural image where resolution enhancement was performed via scaling by a factor of $2$. In the description of the figure, PSNR (in dB), SSIM and S-CIELAB error measure appear in the parenthesis for each method. As can be seen in the enlarged area of Fig. \ref{Fig:Comic2x},  MCcSR more faithfully retains color texture. The bottom row of Fig. \ref{Fig:Comic2x} shows the S-CIELAB error maps for different methods. It is again apparent that the MCcSR method produces less error around edges and color textures. Consistent with the visual observations, the S-CIELAB error is lowest for MCcSR.

Fig. \ref{Fig:Comic3x} also shows the same image with a scaling factor of $3$ and the corresponding S-CIELAB error maps. In this case,  the color texture in the enlarged area is even more pronounced for MCcSR vs. other methods. The trend continues and benefits of MCcSR are most significant for a scaling factor of $4$ in Fig. \ref{Fig:Comic4x}. Similar results for the Baboon image are shown for scaling factors of $2$, $3$, $4$ respectively in Figs. \ref{Fig:Baboon2x}-\ref{Fig:Baboon4x}.


The degradation in image quality for SR results with increased scaling factor is intuitively expected. In a relative sense however, MCcSR suffers a more graceful decay. This is attributed to the use of prior information in the form of the quadratic color regularizers in our cost function, which compensates for the lack of information available to perform the superresolution task.

Tables \ref{Tab:imagesPSNR}-\ref{Tab:imagesSCIELAB} summarize the results of super resolution on images in \emph{set 5} and \emph{set 14} databases with a scaling factor of $3$. PSNR, SSIM and S-CIELAB error measures are compared and almost consistently our MCcSR method outperforms all the other competing state-of-the-art methods. The last row in these tables is essentially the average performance of each method over all the images in {\em set 5} and {\em set 14} datasets. Due to space constraints, we do not include all the LR and SR images for {\em set 5} and {\em set 14} in the paper but they are made  available online in addition to the code at: \tcb{\url{http://signal.ee.psu.edu/MCcSR.html}}.


\begin{figure}
  \centering
  \includegraphics[width=\columnwidth]{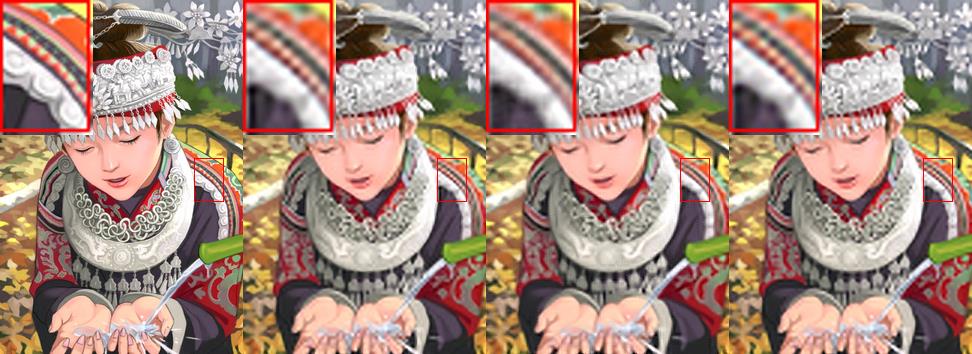}\\\vspace{-0.2in}
  \includegraphics[width=\columnwidth]{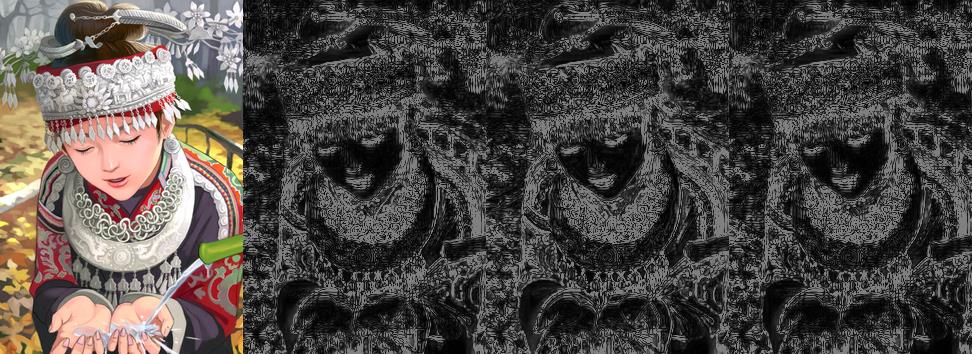}\\\vspace{-0.2in}
  \includegraphics[width=\columnwidth]{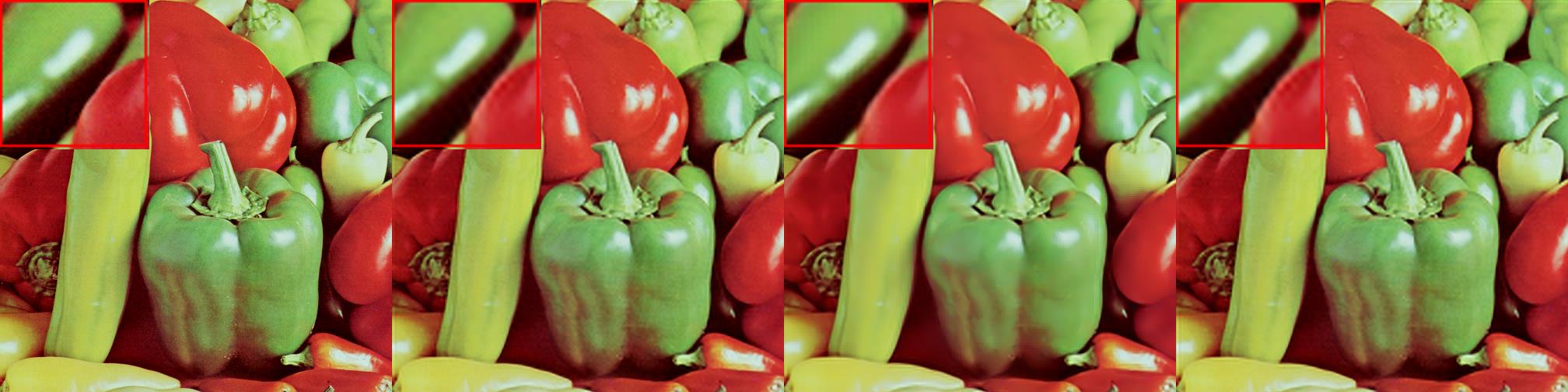}\\\vspace{-0.1in}
  \includegraphics[width=\columnwidth]{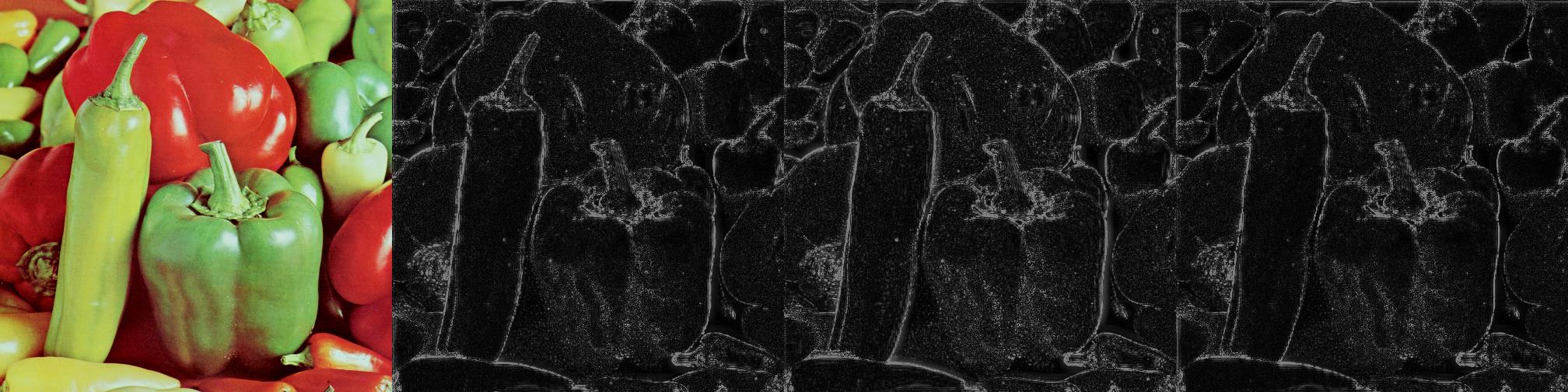}\\\vspace{-0.1in}
  \includegraphics[width=\columnwidth]{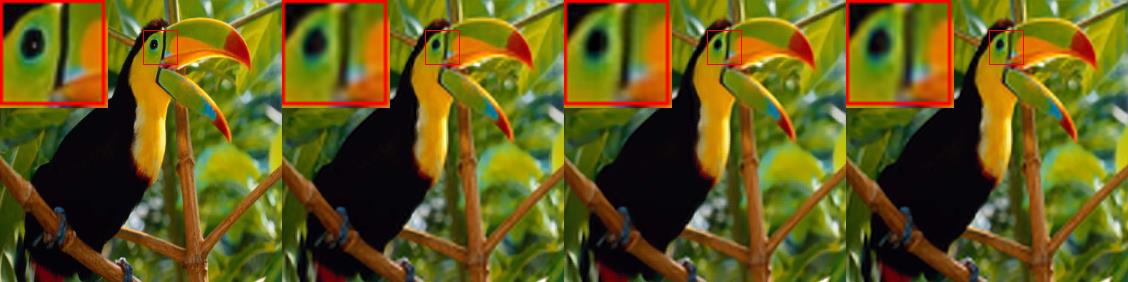}\\
  \includegraphics[width=\columnwidth]{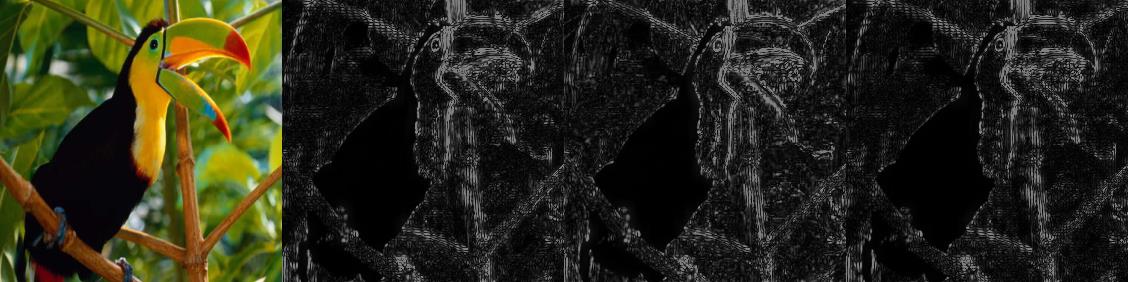}
    \caption{{Visual Images as well as S-CIELAB error maps are shown for a scaling factor of 3. From left to right for each row Images correspond to: Original Image, applying SR separately on RGB channels, ScSR, MCcSR }}\label{Fig:SeparateRGB2}
\end{figure}
\begin{figure*}
  \centering
  \includegraphics[width=\textwidth]{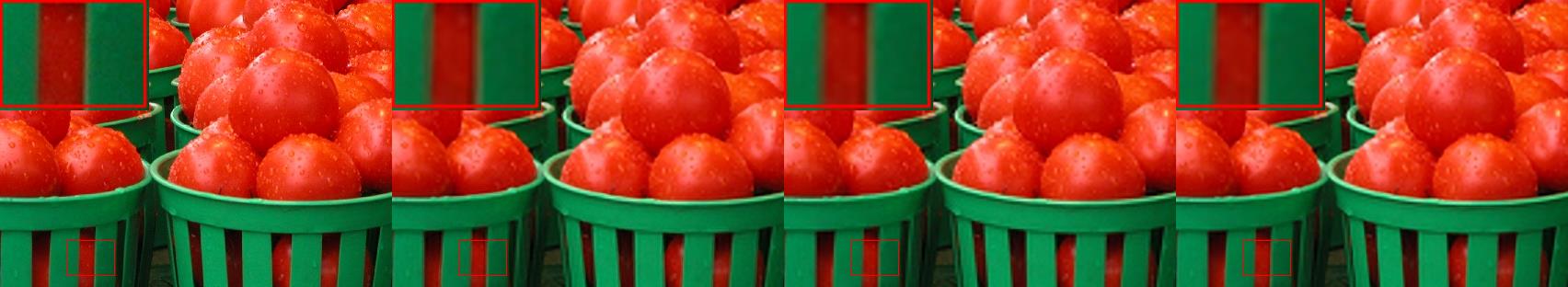}\\
  \includegraphics[width=\textwidth]{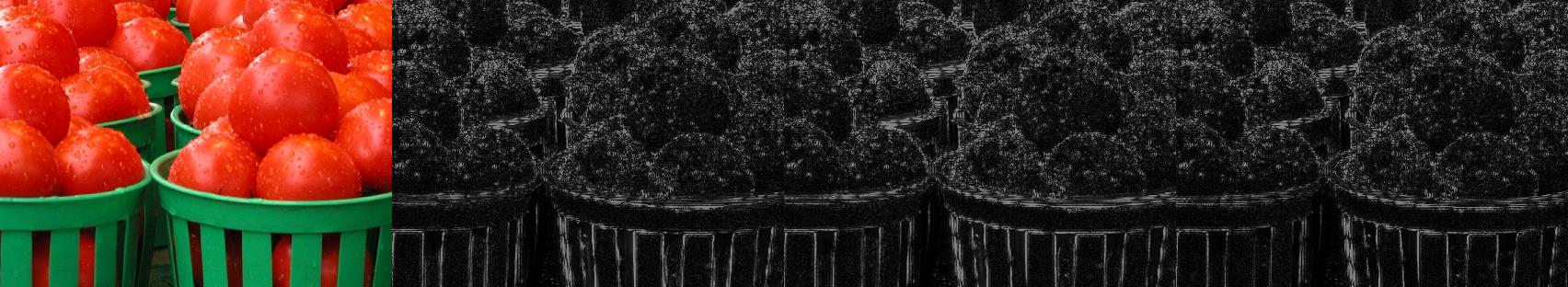}
    \caption{{Visual Images as well as S-CIELAB error maps are shown for a scaling factor of 3. From left to right for each row images correspond to: Original image, applying SR separately on RGB channels (36.26, 0.83, 1.57e4), ScSR (36.13, 0.83, 1.67e4) and \textbf{MCcSR (36.67, 0.85, 1.43e4)}. Numbers in parenthesis are PSNR, SSIM and SCIELAB error measures.}}\label{Fig:SeparateRGB1}
\end{figure*}

\subsection{Effect of Dictionary Size}

So far we have used a fix dictionary of size $512$ atoms for all the methods. In this Section, we evaluate the effect of the learned dictionary size for super-resolution. We again sampled $100,000$ image patches and train $6$ dictionaries of size $16, 32, 64, 128, 256$ and $512$ respectively. The results are evaluated both visually and quantitatively in terms of PSNR, SSIM and S-CIELAB. As is intuitively expected reconstruction artifacts gradually diminish with an increase in dictionary size and our visual observations are also supported by PSNR, SSIM and S-CIELAB of the recovered images. Fig \ref{Fig:DictSizePSNR}  shows the variation of different image quality metrics against dictionary size. For SSIM and S-CIELAB in particular, MCcSR is able to generate effective results even with smaller dictionaries.

\subsection{Effect of Color Regularizers: Separate RGBs}


We provide evidence for the importance of effectively accounting for color geometry via an illustrative example image. Three variations of color SR results are presented next:
\begin{enumerate}
  \item SR performed only on the luminance channel by ScSR \cite{Yang:CoupledDicLearnSR_TIP2012} method and bicubic interpolation is applied for chrominance channels.
  \item Single channel SR performed on red, green and blue channels independently. We again use  ScSR method; however, we learn separate dictionaries for RGB channels and apply ScSR  on RGB channels independently.
  \item Super-resolution  by explicitly incorporating cross channel information into the reconstruction  (our McCSR).
\end{enumerate}
In these experiments we use a scaling factor of $3$ and the results are reported in Figs. \ref{Fig:SeparateRGB1}, \ref{Fig:SeparateRGB2} and Table \ref{Tab:SeparateRGB}. It should particularly be noted (see Fig.\ \ref{Fig:SeparateRGB1}) that applying the SR method independently on RGB channels introduces very significant artifacts around color edges which are not visible in the results of MCcSR and ScSR. Fig.\ \ref{Fig:SeparateRGB2} shows similar results for a few other images. Table \ref{Tab:SeparateRGB} reports image quality measures which confirms the importance of using color channel constraints.


\begin{table*}
\centering
      \caption{Quantitative measures to show effectiveness of color constraints in SR for a scaling factor of 3.}
      \vspace{-0.08in}
      \label{Tab:SeparateRGB}
      \begin{tabular}{ |l||c|c|c|c|c|c|c|c|c| }
      \hline
\multirow{2}{*}{Images}  & \multicolumn{3}{c|}{PSNR (dB)}                     & \multicolumn{3}{c|}{SSIM}                    & \multicolumn{3}{c|}{S-CIELAB} 	\\	
      \cline{2-10}
                        &	Separate RGB      &	ScSR		&	MCcSR	&	Separate RGB      &	ScSR		&	MCcSR	&	Separate RGB      &	ScSR		&	MCcSR		  \\ \hline
comic					&	28.37	          &	28.25	    &\tb{28.51} &	0.74 	          &	0.74 	    &\tb{0.76} 	&	2.80e4            &	3.00e4      &\tb{2.71e4}      \\	\hline
baboon					&	26.95	          &	26.95	    &\tb{27.11}	&	0.53 	          &	0.52 	    &\tb{0.55} 	&	9.93e4            &	1.02e5      &\tb{9.57e4}     	 \\  \hline
pepper					&	36.14	          &	36.20	    &\tb{36.30}	&	0.79 	          &	0.79 	    &\tb{0.80} 	&	1.93e4            &	1.91e4      &\tb{1.73e4}    	 \\  \hline
bird					&	37.71	          &	37.59	    &\tb{38.02}	&	0.92 	          &	0.91	    &\tb{0.93} 	&	7.28e3            &	7.81e3      &\tb{6.50e3}     	 \\  \hline
      \end{tabular}
\end{table*}

\subsection{Robustness to Noise}

An often made assumption in single image SR is that the input images are clean and free of noise which is likely to be violated in many real world applications. Classical methods deal with noisy images by first denoising and filtering out the noise and then performing super-resolution. The final output of such a procedure highly depends on the denoising technique itself and the artifacts introduced in the denoising procedure may remain or even get magnified after super-resolution.

Similar to \cite{YangAndWright:SparseSR_TIP2010}, the parameter $\lambda$ in (\ref{Eq:MainOptProb}) is tuned based on the noise level of the input image and can control the smoothness of output results. We argue that our approach not only benefits from the noise robustness of ScSR \cite{YangAndWright:SparseSR_TIP2010}, but the  additional correlation information from multi-channels can help in further recovering more cleaner images.

\begin{figure}
  \centering
  \includegraphics[width=.98\columnwidth]{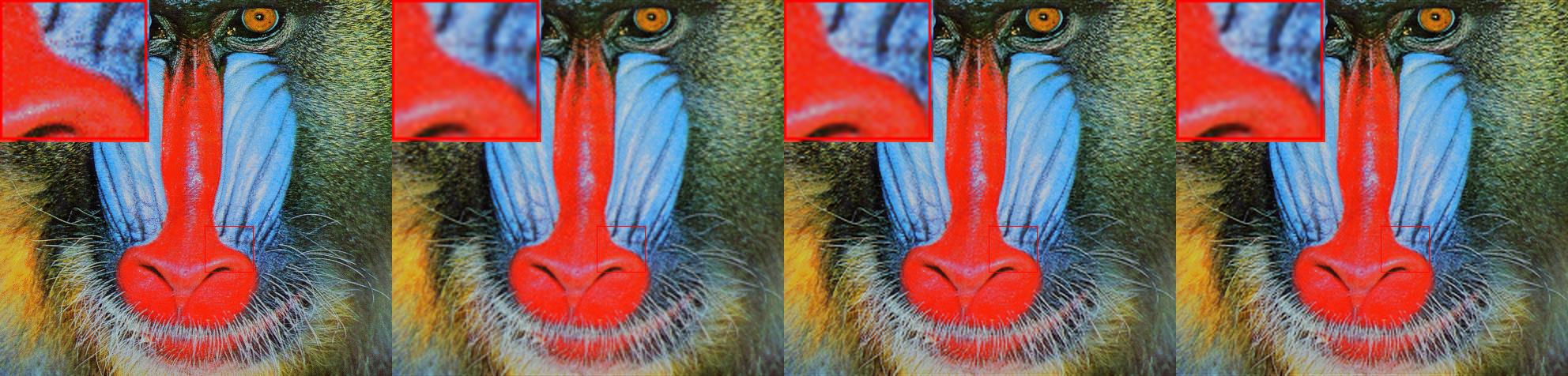}\\
  \includegraphics[width=.98\columnwidth]{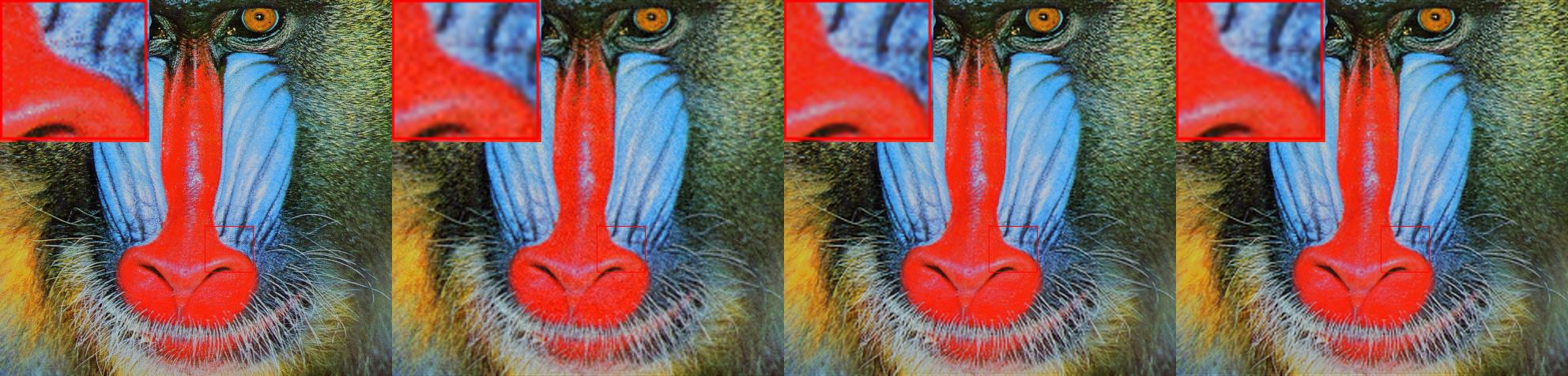}\\
  \includegraphics[width=.98\columnwidth]{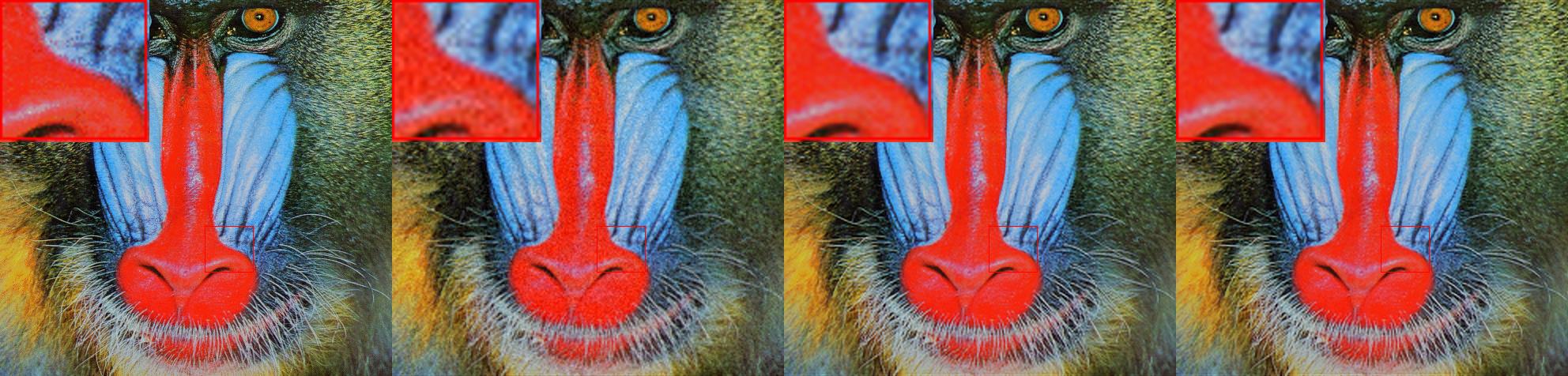}\\
  \includegraphics[width=.98\columnwidth]{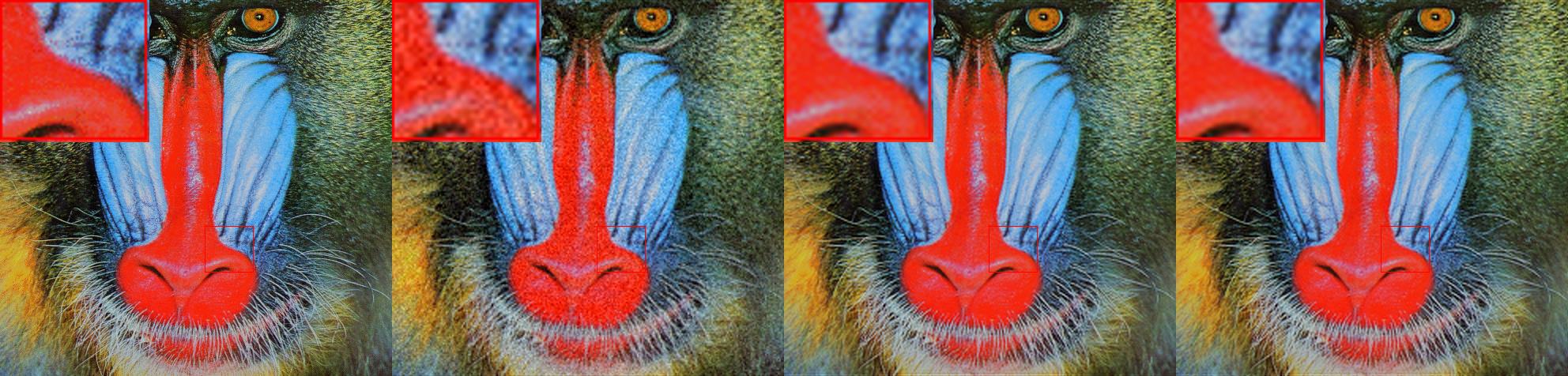}
    \caption{{SR performance under different noise standard deviations: 4,6,8,12 (from top to bottom ) with different methods: Original, bicubic, MCcSR, ScSR (from left to right) }}\label{Fig:NoisyBaboon}
\end{figure}

We add different levels of Gaussian noise to the LR image input to test the robustness of our algorithm to noise and compare our results with ScSR method which has demonstrated success \cite{YangAndWright:SparseSR_TIP2010} in SR in the presence of noise. With a scaling factor of $3$, we chose the range of standard deviation of noise from $4$ to $12$ and similar to \cite{YangAndWright:SparseSR_TIP2010} set $\lambda$ to be one tenth of noise standard deviation. Likewise, we made the choice of $\tau$ in (\ref{Eq:MainOptProb}) using a cross-validation procedure to suppress noise. Fig \ref{Fig:NoisyBaboon} shows the SR results of an image with different levels of noise in comparison with ScSR and bicubic methods.
 Table \ref{Tab:Noise} reports the average PSNR, SSIM and S-CIELAB error measures of reconstructed images from different levels of noisy images. In all cases, MCcSR outperforms the competition.
 \begin{table}
\centering
      \caption{Average performance under different noise levels.}
      \vspace{-0.08in}
      \label{Tab:Noise}
      \begin{tabular}{|l||c|c|c|c|c|c|c| }
\hline
Measure							&	Method                  &   $\sigma=0$ &	$\sigma=4$	&	$\sigma=6$	&	$\sigma=8$	&	$\sigma=12$		 \\  \hline
\multirow{3}{*}{PSNR}			&	Bicubic					&   33.08      &	32.99		&	32.75		&	32.50		&	31.88	 \\	 \cline{2-7}
								&	ScSR					&   34.00      &	33.95		&	33.92		&	33.90		&	33.86	 \\  \cline{2-7}
								&	MCcSR					&   34.14      &	34.11		&	34.09		&	34.09		&	34.07	 \\  \hline
\multirow{3}{*}{SSIM}			&	Bicubic					&   0.745      &	0.731		&	0.698		&	0.672		&	0.619	 \\	 \cline{2-7}
								&	ScSR					&   0.774      &	0.772		&	0.766		&	0.761		&	0.752	 \\	 \cline{2-7}
								&	MCcSR					&   0.785      &	0.783		&	0.780		&	0.775		&	0.768	 \\	 \hline
\multirow{3}{*}{\scriptsize{SCIELAB}}&	Bicubic					&	2.79E4     &	2.92E4  	&	4.40E4		&	5.25E4		&	6.31E4	 \\ \cline{2-7}
								&	ScSR					&	2.28E4	   &	2.31E4		&	2.36E4		&	2.39E4		&	2.43E4	 \\  \cline{2-7}
								&	MCcSR					&	2.14E4	   &	2.16E4		&	2.20E4		&	2.21E4		&	2.23E4	 \\  \hline

      \end{tabular}
\end{table}

\vspace{-0.2in}
\section{Conclusion and Future work}
\label{Sec:Conclusion}
\vspace{-0.05in}
In this work, we extend sparsity based super-resolution to multiple color channels. We demonstrate that by using color information and cross channel constraints, significant improvement over single (luminance) channel sparsity based SR methods can be achieved.
In particular, edge similarities among color bands are exploited as cross channel correlation constraints. These additional constraints lead to new optimization problems both in the sparse coding and learning steps for which we present tractable solutions. Experimental results show the merits of our proposed method both visually and quantitatively.
 While our work offers one possible way to capture cross-channel color constraints, chrominance geometry can be captured via alternative quantitative formulations as in \cite{Srinivas:ColorSR_CIC2011, Farsiu:ColorDemosaicSR_TIP2006, Keren:ColorSR_1999MachineVision, Dai:SoftCutColorSR_2009TIP}. Incorporating these as constraints or regularizers in a sparsity based color SR framework forms a viable direction for future work.

\ifCLASSOPTIONcaptionsoff
  \newpage
\fi



%

\bibliographystyle{IEEEtran}
\bibliography{TIP2016_ver6-Arxiv}

\end{document}